\documentclass[pmlr]{jmlr}

\input{preamble}

\usepackage{amsmath,amsfonts,bm}
\usepackage{mathtools}

\def\eqref#1{equation~\ref{#1}}

\def\1{\bm{1}}

\def\eps{{\epsilon}}

\DeclareMathAlphabet{\mathsfit}{\encodingdefault}{\sfdefault}{m}{sl}
\SetMathAlphabet{\mathsfit}{bold}{\encodingdefault}{\sfdefault}{bx}{n}

\def\gE{{\mathcal{E}}}

\def\gG{{\mathcal{G}}}

\def\gL{{\mathcal{L}}}

\def\gN{{\mathcal{N}}}

\def\gR{{\mathcal{R}}}

\newcommand{\bI}{\mathbf{I}}

\newcommand{\bx}{\mathbf{x}}
\newcommand{\br}{\mathbf{r}}

\begin{document}

\maketitle

\ifarxivorcameraready
    \vspace{-3em}
\fi
\ifdefined\submission
    \vspace{-2em}
\fi
\begin{figure}[!h]
    \centering
    \begin{tabular}{cc|cc}
    Ultra-NeRF & Ours (\acronym) & Ultra-NeRF & Ours (\acronym)\\
    \includegraphics[width=0.23\textwidth]{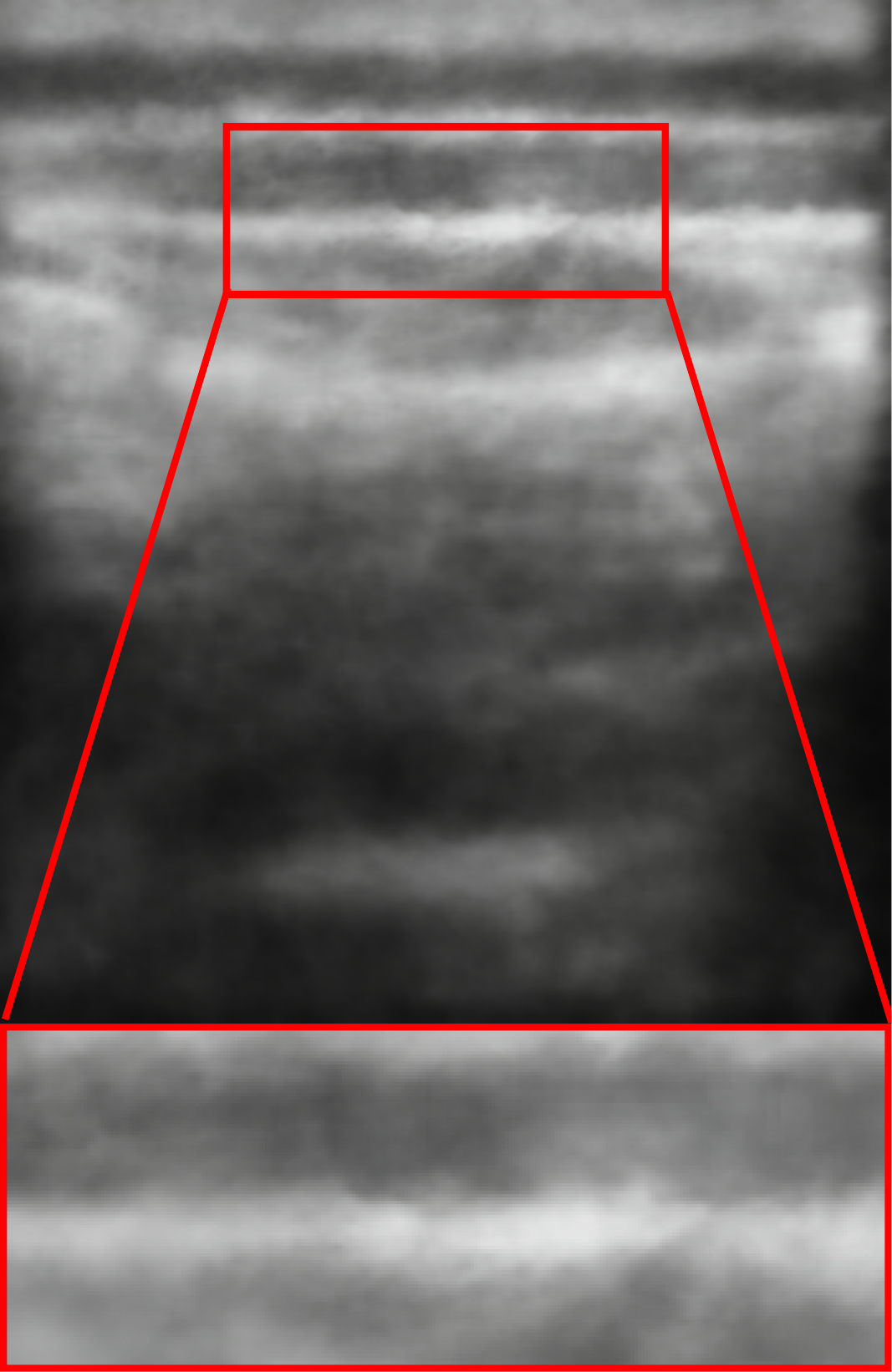} &       \includegraphics[width=0.23\textwidth]{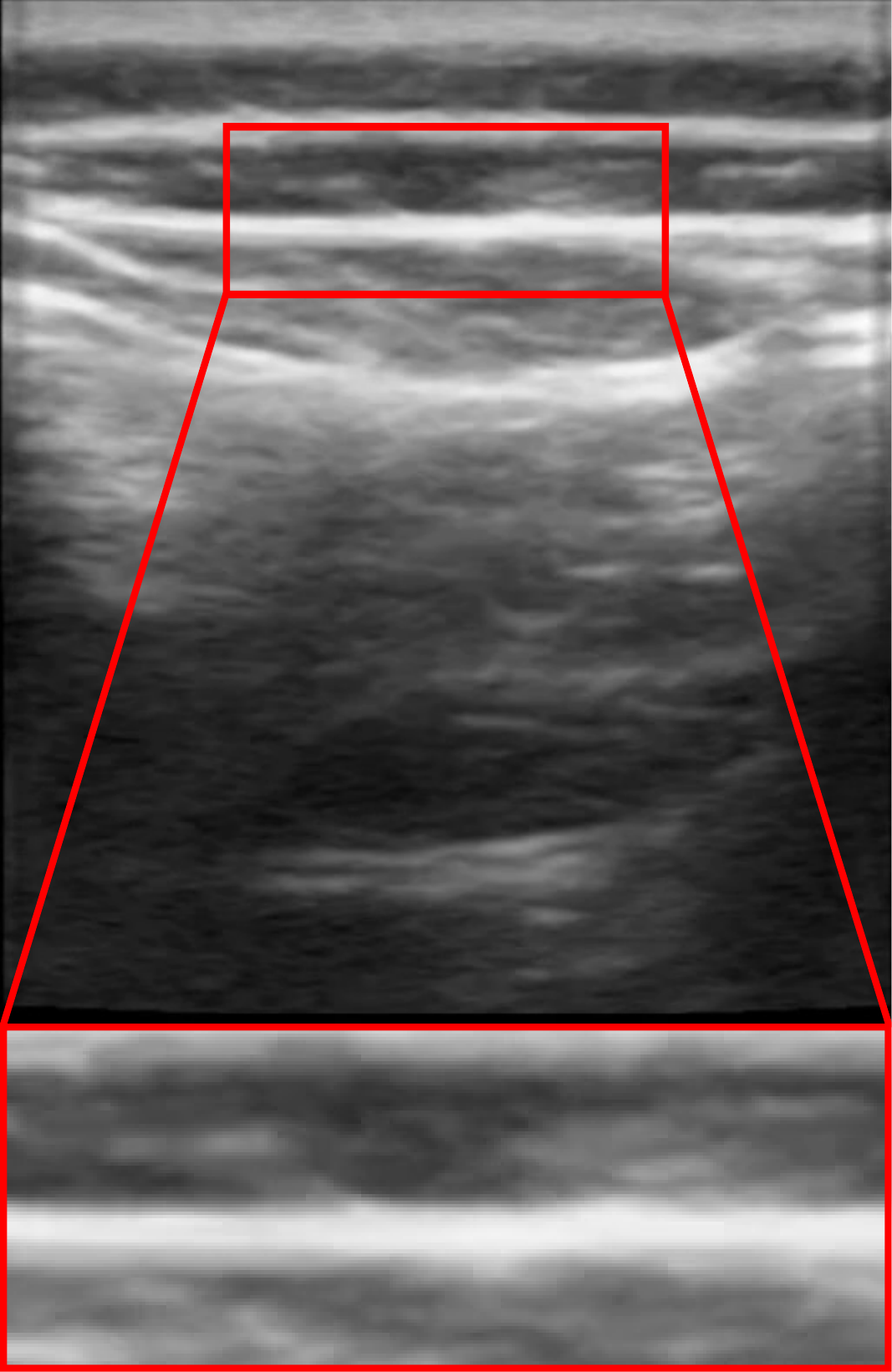} & 
    \includegraphics[width=0.23\textwidth]{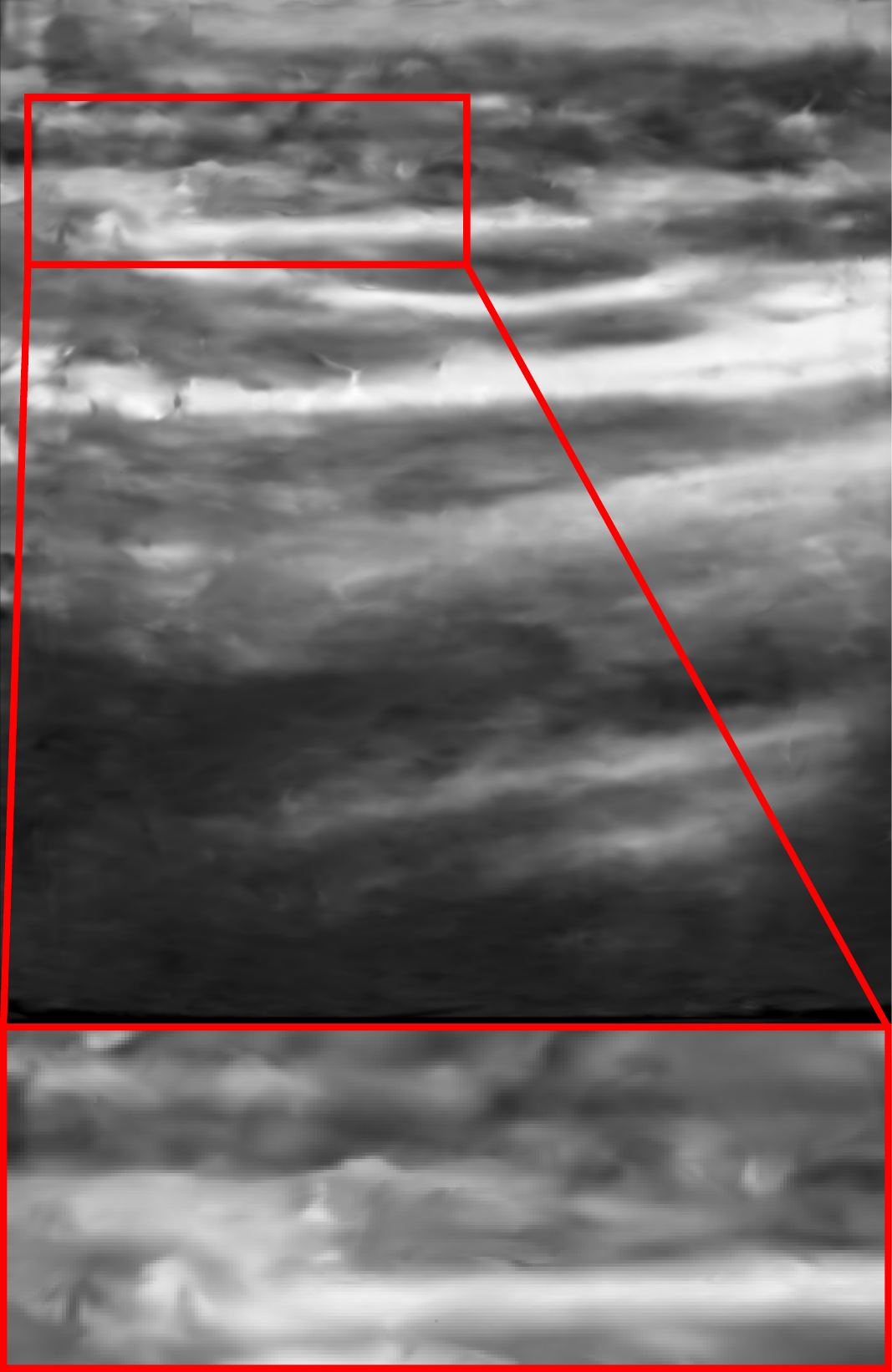} & 
    \includegraphics[width=0.23\textwidth]{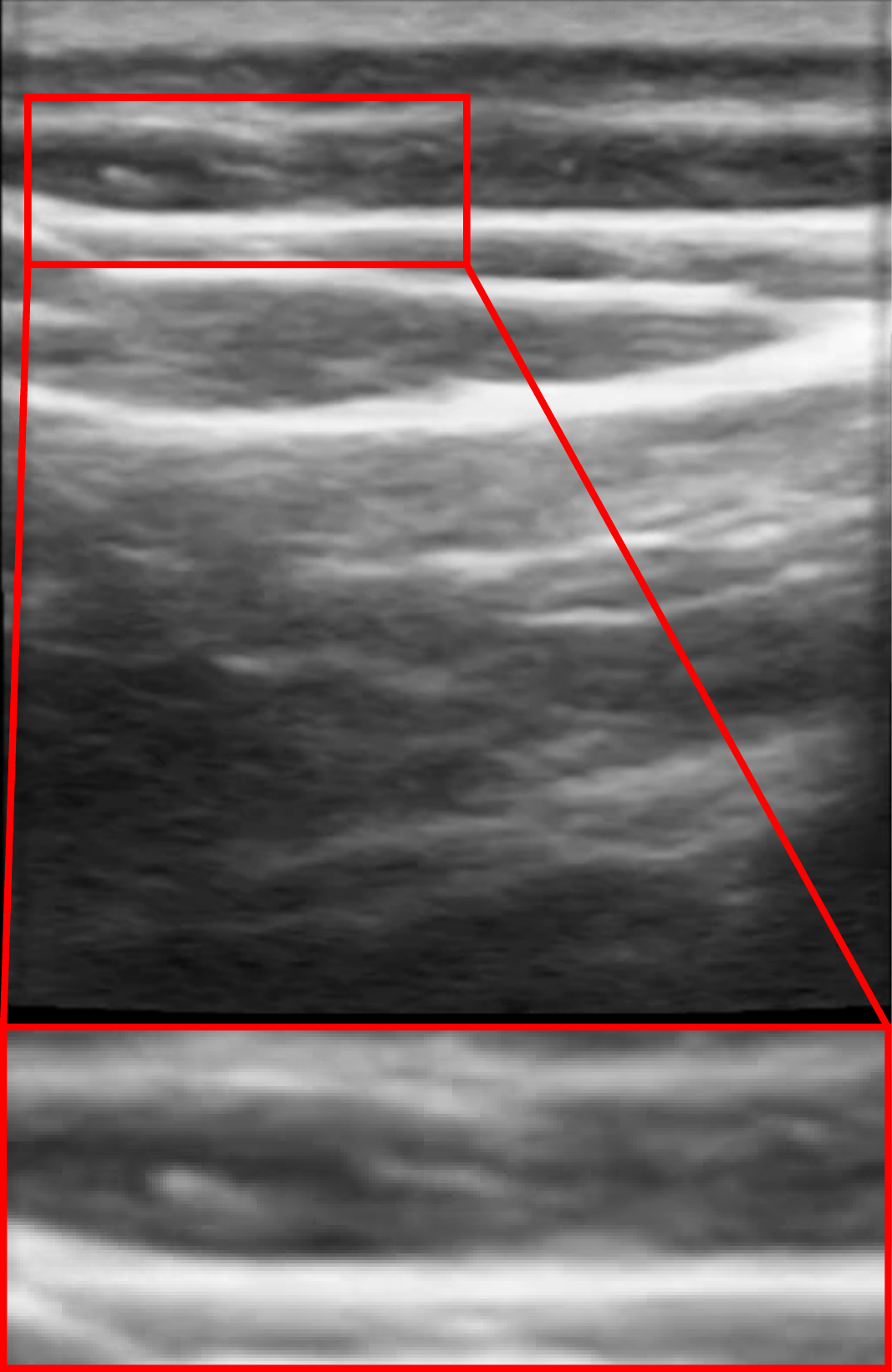} 
    \end{tabular}
    \caption{Comparison of novel ultrasound views rendered by Ultra-NeRF~\citep{pmlr-v227-wysocki24a} (left) and our \acronym\xspace(right). These reconstructions are of the human knee featuring an anteroposterior view (left) and a lateral view (right). Our approach, \acronym\xspace, significantly improves the quality of the reconstructions, making these suitable for any subsequent downstream clinical tasks.}
    \label{fig:teaser}
\end{figure}

\begin{abstract}
Current methods for performing 3D reconstruction and novel view synthesis (NVS) in ultrasound imaging data often face severe artifacts when training NeRF-based approaches. The artifacts produced by current approaches differ from NeRF floaters in general scenes because of the unique nature of ultrasound capture. Furthermore, existing models fail to produce reasonable 3D reconstructions when ultrasound data is captured or obtained casually in uncontrolled environments, which is common in clinical settings. Consequently, existing reconstruction and NVS methods struggle to handle ultrasound motion, fail to capture intricate details, and cannot model transparent and reflective surfaces. In this work, we introduced~\acronym, which incorporates 3D-geometry guidance for border probability and scattering density into NeRF training, while also utilizing ultrasound-specific rendering over traditional volume rendering. These 3D priors are learned through a diffusion model. Through experiments conducted on our new ``Ultrasound in the Wild'' dataset, we observed accurate, clinically plausible, artifact-free reconstructions.
\end{abstract}

\section{Introduction}

\begin{figure}[tb]
    \centering
    \includegraphics{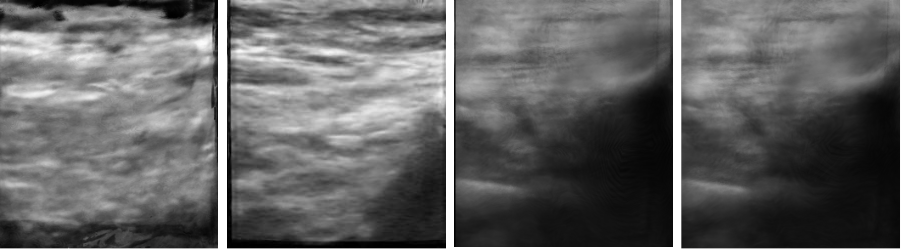}
    \caption{We observe the existence of multiple NeRF artifacts in ultrasound imaging which is a common challenge in medical NeRF-based methods.}
    \label{fig:usfloater}
\end{figure}

All imaging systems benefit from capturing the 3D geometry of the scenes being imaged. Capturing the 3D geometry of scenes is crucial in medical imaging, not only for accurate diagnosis but also for effective treatment planning. This is particularly vital in conditions like hemophilic arthropathy, where joint bleeds lead to synovial proliferation and effusion, and the measurement of synovial fluid volume is a common diagnostic step. Among the key indicators of disease activity in the joints is synovial recess distention (SRD), which may arise from various factors such as accumulation of blood or synovial fluid~\citep{heilmann1996synovial}. Similarly, diagnosing conditions such as scoliosis requires understanding the volumetric properties of the spine~\citep{kadoury2007three,4804738}. Producing robust and accurate 3D representations while performing view synthesis from 2D medical imaging techniques is an important problem that can be used for planning and downstream clinical tasks~\citep{doi:10.1177/1941738112468416,jackson1988magnetic,ji2011real}.

We are interested in the problem of 3D reconstruction and novel view synthesis for ultrasound imaging. We focus our efforts on working with ultrasound imaging because it is one of the most cost-effective and accessible forms of medical imaging. We are interested in performing this task on casually captured or in the wild ultrasounds since all ultrasounds captured by clinicians are casually captured. Furthermore, performing this task with ultrasound imaging is also a very challenging task due to the nature of how ultrasounds are captured. This task has traditionally been performed by training clinicians to mentally assemble a 3D image. There have been some digital techniques towards the creation of such 3D ultrasound images which relied on utilizing advanced wobbler probes~\citep{morgan2018versatile}, 2D transducers~\citep{smith2002two}, or tracking probes~\citep{POON20051095}. These kinds of approaches are often based upon assembling a 3D image from multiple 2D slices and use a lot of handcrafted priors. Thus, these classes of approaches often have many limitations, primarily these approaches are often unable to produce plausible 3D reconstructions that could be used for downstream tasks~\citep{kojcev2017reproducibility}. Such manual and digital approaches end up being very costly, error-prone, and irreproducible~\citep{kojcev2017reproducibility,lyshchik2004three}.

Some recently learned digital approaches based on neural radiance field~\citep{10.1145/3503250} methods have also been developed that work toward the problem we highlight~\citep{pmlr-v227-wysocki24a,gaits:hal-04480668,9593917}. These kinds of approaches have also successfully been used in other medical contexts like reconstructing CT projections from X-ray~\citep{coronafigueroa2022mednerf} and surgical scene 3D reconstruction~\citep{wang2022neural,10.1007/978-3-031-43996-4_2} which have shown impressive results. However, using such methods for ultrasound imaging does not produce artifact-free 3D representations from casually captured ultrasounds.

\begin{figure}[tb!]
    \vspace{0em}
    \centering
    \floatbox[{\capbeside\thisfloatsetup{capbesideposition={right,top},capbesidewidth=9.15855cm}}]{figure}[\FBwidth]
    {\caption{\protect\vspace{-1em} We show a depth map of two reconstructions produced by our approach, NeRF-US. We superimpose the depth map with tissue boundaries as shown by color-coded \colorbox{vlet}{curves}. Our approach produces accurate representations with the tissue boundaries unambiguously reconstructed.}\label{fig:param}}
    {\includegraphics[width=6.07927cm]{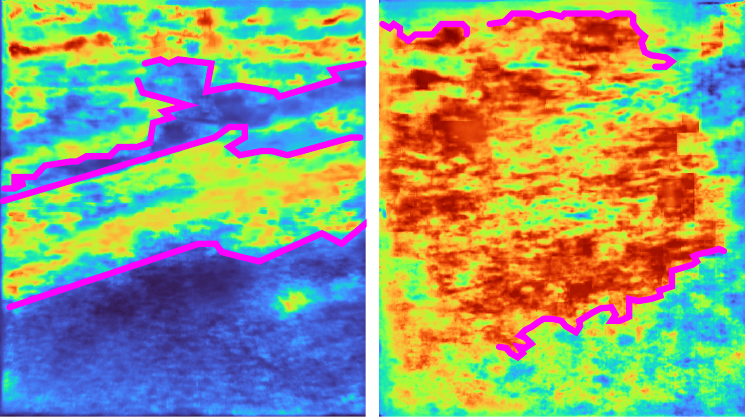}}
\end{figure}

There are a few challenges common across all these learned digital methods: the need for high-quality diverse datasets that capture intricate details like tissue interface locations in high detail, accurately modeling transparent and reflective surfaces, specular surface rendering, and delineating boundaries between different tissues. We qualitatively show some NeRF artifacts that appear when Ultra-NeRF is adopted~\citep{pmlr-v227-wysocki24a} on ultrasound images in the wild in~\Cref{fig:usfloater}. These challenges are common across all medical NeRF-based methods. In contrast, our approach, \acronym, produces artifact-free reconstructions with minor details that are accurately reconstructed, as shown in~\Cref{fig:param}.

Towards the problem of producing high-quality ultrasound reconstructions in the 'wild', we propose our approach, \acronym. Our goal is to produce a 3D representation given a set of ultrasound images taken in the wild and their estimated camera or ultrasound probe positions. We first train a 3D denoising diffusion model which serves as geometric priors for the reconstruction. We then train a NeRF model that takes in a 3D vector (denoting positions in 3D) and learns a 5D vector (attenuation, reflectance, border probability, scattering density, and scattering intensity) following the success of Ultra-NeRF~\citep{pmlr-v227-wysocki24a}. While training this NeRF, we incorporate the geometric priors from the diffusion model to guide the outputs for border probability and scattering density. This allows our approach, \acronym, to accurately generate 3D representations for ultrasound imaging in the 'wild', as shown in~\Cref{fig:teaser}. We also evaluate our approach with a new dataset for this task, which we will release publicly. To the best of our knowledge, ours is the first approach that handles this challenging task of reconstructing noniconic, nonideal ultrasound images in the 'wild' and also outperforms previous methods.

Our approach is universal in ultrasound imaging and our experiments include human spine and knee joint ultrasound. We observe that current evaluation and management strategies for painful musculoskeletal episodes in patients with hemophilia often rely on subjective evaluations, leading to discrepancies between patient-reported symptoms and imaging findings. Ultrasonography (US) has emerged as a valuable tool for assessing joint health due to its accessibility, safety, and ability to detect soft tissue changes, including joint recess distension, with precision comparable to magnetic resonance imaging (MRI). However, accurately interpreting US images, especially in identifying and quantifying recess distention, remains challenging for clinicians.
Providing clinicians with a 3D representation of recess distension acquired from US cine loops could greatly enhance their ability to appreciate the extent of fluid accumulation within the joint space. This information is crucial for guiding treatment decisions, especially regarding the initiation and optimization of hemostatic therapy and the utilization of adjunctive treatments such as physical therapy and anti-inflammatory agents. Thus, our new ``Ultrasound in the wild'' dataset is focused on ultrasounds around the knee for the recess distention problem.

\ifdefined\cameraready
\subsection*{Generalizable Insights about Machine Learning in the Context of Healthcare}
\fi

Addressing the challenge of casually captured ultrasounds, which is the common practice among clinicians today, \acronym integrates specific ultrasound properties within the NeRF framework. This integration not only surpasses existing methods in producing artifact-free 3D reconstructions, but also expands the potential applications beyond routine medical diagnostic tasks. For example, the ability to generate accurate and artifact-free 3D models from ultrasound data can significantly enhance surgical planning and postoperative evaluations. In addition, our approach aims to increase the use of cost-effective methods such as ultrasound, offering a viable alternative to more expensive imaging techniques for certain medical scenarios.

\paragraph{Contributions.} Modern 3D reconstruction methods for ultrasound imaging are either overly reliant on hand-crafted priors, manual intervention, or are unable to work on ultrasound imaging in the `wild'. The key novelty of our approach stems from the modification of NeRF-based methods for this task.
\begin{itemize}
    \item We propose a first-of-its-kind approach for training NeRFs on ultrasound imaging that incorporates the properties of ultrasound imaging and incorporates 3D priors through a diffusion model to reconstruct accurate images in uncontrolled environments.
    \item We introduce a new dataset, ``Ultrasound in the Wild'', featuring real, non-iconic ultrasound imaging of the human knee. This dataset will serve as a resource for benchmarking NVS performance on ultrasound imaging under real-world conditions and will be made publicly available.
    \item Our approach demonstrates improved qualitative and quantitative performance by significantly reducing or eliminating ultrasound imaging artifacts. This leads to more accurate 3D reconstructions compared to other medical and non-medical NeRF-based methods. Furthermore, we observe clinically plausible 3D reconstructions from ultrasound imaging in the `wild'.
\end{itemize}
We open-source our code and data on our project webpage.

\section{Related Works}

\paragraph{Implicit Neural Representations and View Synthesis.} Historically, the field of novel view synthesis relied on traditional techniques like image interpolation~\citep{10.1145/2487228.2487238,10.1145/3596711.3596757,10.1145/166117.166153,10.1145/3596711.3596761,10.1145/237170.237191,fitzgibbon2005image,seitz1996view,10.1145/3596711.3596758,10.1145/218380.218398} and light field manipulation~\citep{sloan1997time,10.1145/237170.237199} to generate new views without needing to understand the geometry of the scene. These methods worked best with densely sampled scenes, which limited their usability. The advent of learned techniques, popularized using image blending~\citep{Flynn_2016_CVPR,10.1145/3272127.3275084} and multiplane images~\citep{10.1007/978-3-030-58452-8_11,10.1145/3306346.3322980}. Further progress involved creating explicit 3D scene representations through meshes~\citep{10.1007/978-3-7091-6453-2_10,10.1007/978-3-030-58529-7_37,6599051}, point clouds~\citep{qi2017pointnet,10.1007/978-3-030-58542-6_42,bui2018point,10.1007/978-3-7091-6453-2_17}, and voxel grids~\citep{791235,10.1145/3130800.3130855,seitz1999photorealistic}, offering a better understanding of scene geometry but with many challenges in model accuracy and robustness.

The introduction of neural implicit representations (INRs) marked a change in representing 3D scenes, offering a more adaptable and comprehensive method for encoding both the geometric and appearance aspects of scenes~\citep{sitzmann2019scene,sitzmann2020implicit}. Utilizing neural networks as a foundation, this approach allowed for the implicit modeling of scene geometry, with the capacity to generate novel views through ray-tracing techniques. By interpreting a scene as a continuous neural-based function, INRs map 3D coordinates to color intensity. Among the diverse neural representations for 3D reconstruction and rendering, the success of deep learning has popularized methods like Level Set-based representations, which map spatial coordinates to a signed distance function (SDF)~\citep{Park_2019_CVPR,Jiang_2020_CVPR,yariv2020multiview} or occupancy fields~\citep{Mescheder_2019_CVPR}.

One of the most popular INR methods is the Neural Radiance Field (NeRF)~\citep{10.1145/3503250}, an alternative implicit representation that directly maps spatial coordinates and viewing angles to local point radiance. NeRFs can also perfrom novel view synthesis through differentiable volumetric rendering, utilizing only RGB supervision and known camera poses. Furthermore, there have also been many extensions of NeRF-based methods for dynamic scenes~\citep{Pumarola20arxiv_D_NeRF,Gafni20arxiv_DNRF,Park20arxiv_nerfies}, compression~\citep{takikawa2022variable}, editing~\citep{liu2021editing,zhang2021stnerf,liu2022nerfin}, and more.

\begin{figure}[tb]
    \centering
    \includegraphics{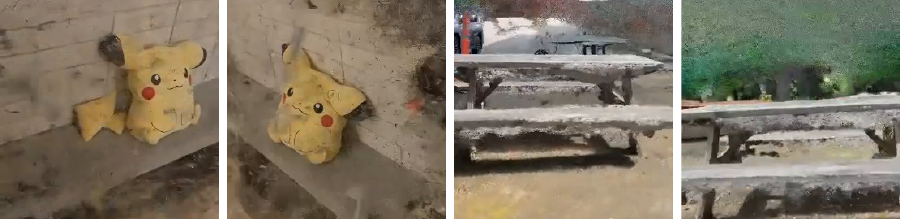}
    \caption{Floater artifacts often appear in NeRF-based models for casually-captured videos.}
    \label{fig:floater}
\end{figure}

\paragraph{NeRFs in the wild.} Applying NeRFs in uncontrolled environments often results in the emergence of artifacts, often referred to as \emph{floaters}. These are small, disconnected regions in volumetric space that inaccurately represent parts of the scene when viewed from different angles, appearing as blurry clouds or distortions in the scene which we demonstrate in~\Cref{fig:floater}. These artifacts are primarily observed under conditions such as sub-optimal camera registration~\citep{Nerfbusters2023}, sparse input sets~\citep{liu2023cleannerf,Nerfbusters2023}, strong view-dependent effects~\citep{liu2023cleannerf}, and inaccuracies in scene geometry estimation. Additionally, the very nature of NeRF's approach, which reconstructs scene geometry and texture from 2D projections of a 3D scene causes information loss.

To mitigate floater artifacts in NeRF reconstructions, several strategies focus on improving camera pose estimation and scene geometry. Mip-NeRF 360~\citep{Barron_2022_CVPR} and RegNeRF~\citep{Niemeyer_2022_CVPR} enhance alignment and consistency, with the former also incorporating a distortion loss to prevent floaters. Techniques such as Nerfbusters~\citep{Nerfbusters2023} and ViP-NeRF~\citep{10.1145/3588432.3591539} leverage visibility information to better reconstruct scenes with sparse data, while CleanNeRF~\citep{liu2023cleannerf} offers a method to separate view-dependent effects for more accurate geometry estimation. Additionally, previous works have also explored post-processing methods to remove floaters without modifying the original NeRF framework~\citep{jambon2023nerfshop,wirth2023post,goli2023bayes}.



\paragraph{NeRFs for Medical Imaging.} The application of NeRFs in medical imaging is confronted with unique challenges that stem from the inherent properties of medical imaging data and the demand for accurate representations. These include the need for detailed inner structure representation, the handling of ambiguous object boundaries, the significance of color density variations in medical images, and the adaptation to different imaging principles compared to traditional NeRF applications~\citep{wang2024neural}.

Recent advances have demonstrated that NeRF-based techniques, such as MedNeRF~\citep{9871757} and UMedNeRF~\citep{hu2024umednerf}, as effective in reconstructing CT projections from single X-ray views, showcasing their potential to minimize exposure to ionizing radiation in medical imaging. ACNeRF~\citep{sun2024acnerf} further enhances reconstruction quality through improved alignment and pose correction. 


There have also been other specialized NeRFs either for a particular body part or a kind of imaging techique. In brain imaging, advances such as 3D reconstruction from MRI scans using NeRFs~\citep{iddrisu20233d} aim to improve diagnostics and patient care by providing more precise and less invasive diagnostic methods. Similarly, in dental and maxillofacial imaging, Masked NeRF~\citep{zhou2023robust} has been introduced to address challenges related to skull CBCT reconstructions, offering refined methods to mitigate artifacts and improve pose estimation accuracy. For cardiovascular imaging, there has been work on 3D reconstruction of coronary angiography images~\citep{zha2022naf,maas2023nerf} and improved diagnostic capabilities. Similarly, NeRF-based models have been developed for feet~\citep{zha2022naf}, chest CT imaging~\citep{9871757,hu2024umednerf,sun2024acnerf,maas2023nerf}, and abdominal surgical planning~\citep{wang2022neural}.

Furthermore, there has been similar work to ours, leveraging ultrasounds to perform 3D reconstruction. Particularly, there have been a few studies performing 3D reconstruction on ultrasound images~\citep{li20213d,yeung2021implicitvol,gu2022representing,song2022development}. Ultra-NeRF~\citep{pmlr-v227-wysocki24a} is another popular technique that encodes the physics of ultrasound into volume rendering and demonstrates promising results for liver and spine ultrasounds. However, such methods still pose multiple challenges in adopting NeRF-based methods for ultrasounds. These methods qualitatively produce multiple ultrasound imaging artifacts, are unable to capture intricate details like tissue boundaries and all ultrasound imaging performed by clinicians inherently has some body motion in it which these methods do not account for. These limit such methods from being applied in the wild. However, our method, \acronym, tackles these challenges common across all current studies.

\section{Preliminary}

\subsection{Diffusion Models}
\label{sec:backgrounddiffusion}

The denoising diffusion probabilistic model (DDPM)~\citep{ho2020denoising} in the forward process takes an input image $\bx_0 \sim q(\bx_0)$, and progressively in the $T$ steps add Gaussian noise to the image. This is implemented using a Markov chain of $T$ steps.
\begin{equation}
\begin{split}
    q(\bx_t \mid \bx_{t-1}) &= \gN(\bx_t; \mu_t = \sqrt{1-\beta_t}\cdot\bx_{t-1}, \beta_t\bI) \\
    &= \gN(\bx_t; \sqrt{\Bar{\alpha_t}}\bx_0, (1-\Bar{\alpha_t})\bI)
\end{split}
\end{equation}

At each step of the Markov chain, the forward process adds a Gaussian noise with variance $\beta_t$ to $\bx_{t-1}$, $\Bar{\alpha_t} = \prod_{s=0}^t 1-\beta_s$, $\beta_t$ is a hyper-parameter representing a variance schedule, and produces a new latent variable $\bx_t$. The reverse process of diffusion models aims to recover the data distribution from the Gaussian noises by approximating the posterior distribution $q(\bx_{t-1}\mid\bx_t, \bx_0)$ as,
\begin{equation}
    p_{\theta}(\bx_{t-1} | \bx_t) = \gN(\bx_{t-1}; \mu_{\theta}(\bx_t, t), \Sigma_{\theta}(\bx_t, t))
\end{equation}

This only requires approximating the mean $\mu_\theta(\bx_t, t)$ by training some neural network $\epsilon_\theta(\bx_t, t)$, this network can be trained using the optimization objective,
\begin{equation}
    \mathcal{L}_{t} = \gE_{\bx_0, t, \epsilon} \left[ \left\lVert \epsilon - \theta(\sqrt{\bar{\alpha}_t \bx_0} + \sqrt{1 - \bar{\alpha}_t \epsilon}, t) \right\rVert^2 \right]
\end{equation}

\subsection{Neural Radiance Fields}
\label{sec:backgroundnerf}

Neural Radiance Fields (NeRFs)~\citep{10.1145/3503250} use a single 5-D coordinate as input $(x, y, z, \theta, \phi)$ representing the spatial location and viewing angle, and outputs $(r, g, b, \sigma)$ representing color intensities and volume density. NeRFs usually output different representations for the same point when viewed from different camera angles which allows us to capture various lighting effects as well. To train these networks without ground truth density and color, we sample pixels from the original images, using ray marching. For a given pixel we have the ray,
\begin{equation}
    r(t) = o+td
\end{equation}
where $o$ represents the origin and $d$ the direction, which we can sample at timesteps $t$. To map these back to an image we can integrate these rays (differentiable rendering),
\begin{equation}
    C(r) = \int_{t_n}^{t_f} \underbrace{T(t)}_{\text{transmittance}}\cdot\overbrace{\sigma(r(t))}^{\text{density}} \cdot \underbrace{c(r(t), d)}_{\text{color}} \ dt
\label{eq:vol}
\end{equation}

We can now train the network by simply computing $L_2$ loss since from~\Cref{eq:vol} we have a way to map the neural field output back to a 2D image.

\section{\acronym}
\label{sec:methods}

\begin{figure}[tb]
    \centering
    \includegraphics[width=\textwidth]{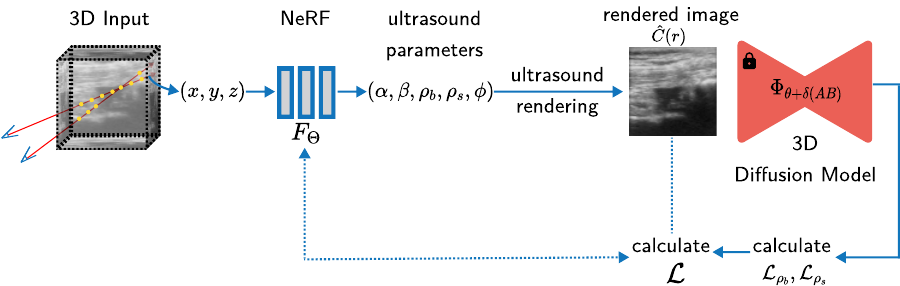}
    \caption{\textbf{An overview of how our method works.} We train a NeRF model (\Cref{sec:methodsrendering}) that uses ultrasound rendering to convert the representations into a 2D image after which we infer through a 3D diffusion model (\Cref{sec:methodsdiffusion}) which has geometry priors through which we calculate a modified loss definition to train the NeRF.}
    \label{fig:methodsoverview}
\end{figure}

Our goal is to produce a 3D representation given a set of ultrasound images taken in the 'wild' and their camera positions. Our approach on a high level involves modifying volumetric rendering while training NeRFs, and using a Diffusion Model to produce clean artifact-free representations as summarized in~\Cref{fig:methodsoverview}. We show our approach for training the diffusion model (\Cref{sec:methodsdiffusion}) and our approach on training the NeRF model (\Cref{sec:methodsrendering}). Notice that, while our training approaches utilize data around the human knee, our work is not limited to this single application. It can be extended right out-of-the-box to any kind of medical ultrasound imaging. We limit our experiments to just the human knee to be able to collect sufficient data and to be able to evaluate our approach satisfactorily.

\subsection{Training the Diffusion Model}
\label{sec:methodsdiffusion}

The first step of our approach relies on the training of a 3D diffusion model, which can serve as geometric priors for our NeRF model (\Cref{sec:methodsrendering}).

Similar to Nerfbusters~\citep{Nerfbusters2023}, this diffusion model produces an $32\times 32\times 32$ occupancy grid, $x$. We use the 3D diffusion model $\Phi$ with parameters $\theta$ trained on ShapeNet~\citep{chang2015shapenet}, from Nerfbusters~\citep{Nerfbusters2023}. Based on LoRA~\citep{hu2022lora}, 
our fine-tuning process now utilizes the same loss function as the original DDPM training, with the low-rank update applied to the model's parameters,
\begin{equation}
\label{eq:ft}
    \gL_{\text{FT}} = \left\| x - \Phi_{\theta+\delta(AB)}( \bar{x}_t + (1-\bar{\beta}_t), t) \right\|_2^2 
\end{equation}
where $x$ is the target occupancy grid, $\bar{x}_t$ is the noised version of $x$ at timestep $t$, $\delta$ is a scaling factor determining the magnitude of adaptation and $AB$ represents a low-rank update to the original weights, and $\bar{\beta}_t$ controls the noise level as per the noise schedule. The model $\Phi_{\theta + \delta(AB)}$ indicates the fine-tuned model. We make sure to only update the parameters of $A$ and $B$ matrices are updated during this process, leaving the original model parameters $\theta$ fixed.

\begin{figure}[tb]
    \centering
    \includegraphics[width=\textwidth]{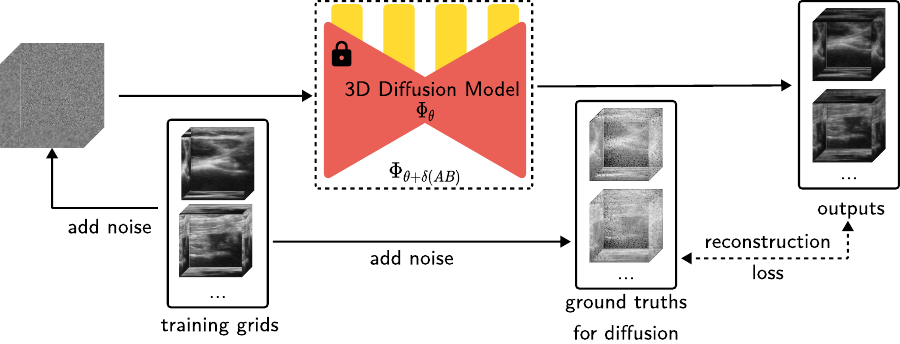}
    \caption{\textbf{Training our Diffusion Model.} An overview of how our diffusion model is fine-tuned, we use $32^3$-sized patches to LoRA-finetune a 3D diffusion model trained on ShapeNet~\citep{chang2015shapenet}.}
    \label{fig:methodsdiffusion}
\end{figure}

We finetune the 3D diffusion model on a small dataset of voxels around the human knee generated synthetically. From these synthetic knees, we extracted localized, cube-bounded patches that represent the area of interest for ultrasound imaging. We particularly ensure that all the voxels we sample have one of their ends on human skin mimicking ultrasound imaging conditions. These cubes are variably sized proportional to the knee model's bounding volume and are voxelized and scaled. We finetune the model $\Phi$ as summarized in~\Cref{fig:methodsdiffusion}, on these $32\times 32\times 32$-sized voxels to produce an adapted model $\Phi_{\theta+\delta(AB)}$ which we use as our 3D geometric priors.

\subsection{Training the NeRF}
\label{sec:methodsrendering}

Following previous ultrasound represenations~\citep{Salehi2015Patientspecific3U,pmlr-v227-wysocki24a}, the parameter vector for some position $q = (x, y, z)$ is $\left[\alpha(q), \beta(q), \rho_b(q), \rho_s(q), \phi(q)\right]$ where $\alpha$ is the attenuation, $\beta$ is the reflectance, $\rho_b$ is border probability, $\rho_s$ is the scaterring density, and $\phi$ is the scaterring intensity. Just like standard NeRF models, we employ an MLP to learn the mapping,
\begin{equation*}
    \mathbf{F}_\Theta: q \rightarrow \left[\alpha, \beta, \rho_b, \rho_s, \phi\right]
\end{equation*}

Ultra-NeRF~\citep{pmlr-v227-wysocki24a} for the volume rendering defines the intensity as
\begin{equation}
\label{eq:unerfvol}
    I(r, t)=I_0 \cdot \prod_{n=0}^{t-1}[(1-\beta(r, n)) \cdot G(r, n)] \cdot \exp \left(-\int_{n=0}^{t-1}(\alpha \cdot f \cdot d t)\right),
\end{equation}
where $\beta(r, t)$ represents the reflection coefficient, $I_0$ represents the initial unit intensity, $G(r, t)$ represents a boundary mask, and $\left(\alpha f d t\right)$ represents the loss of energy due to attenuation at each step of the propagation with $f$ as the frequency. Composing these intensities a 2D ultrasound image is generated which can be trained using the standard NeRF techniques.

We particularly modify this training process inspired by the area of work that incorporate diffusion models in the NeRF training~\citep{wynn-2023-diffusionerf,Nerfbusters2023,single-stage-diffusion,nerfdiff,yang2023learning}, however incorporating these with the rendering process is not straightforward. We do so leveraging the diffusion model we trained $\Phi_{\theta+\delta(AB)}$ to refine key ultrasound parameters: border probability $\rho_b$, scattering density $\rho_s$ for which we can efficiently use the 3D geometric priors. Given a set of ultrasound parameters $\gG_i = \{\rho_b, \rho_s\}$ for each voxel $i$, we process these through the diffusion model to generate predictions on the expected values of these parameters. We then denote the two new loss terms as,
\begin{equation}
\label{eq:lossterms}
    \begin{split}
        \gL_{\rho_b} &= \frac{1}{N}\sum_{i=1}^N \left\lvert \gG_i^{\rho_b} - m_i^{\rho_b} \right\rvert^2 \\
        \gL_{\rho_s} &= \frac{1}{N}\sum_{i=1}^N \left\lvert \gG_i^{\rho_s} - m_i^{\rho_s} \right\rvert^2 , \\
    \end{split}
\end{equation}
where $\rho_b$ represents the border probability, $\rho_s$ represents the scattering density, $N$ represents the number of points being sampled, $m_i^{\rho_b}$ represents the value of $\rho_b$ from the output of the diffusion model $\Phi_{\theta+\delta(AB)}$ for a voxel $i$, and $m_i^{\rho_s}$ represents the value of $\rho_s$ from the output of the diffusion model $\Phi_{\theta+\delta(AB)}$ for a voxel $i$. We find that formulating the two loss components as shown in~\Cref{eq:lossterms} to perform particularly well since many of the underlying objects captured by ultrasound imaging can often contain opaque, translucent, and transparent in the same scene (note: that our voxels are rather small and often only contain a single kind of object).

We now write our final loss definition for the NeRF training as
\begin{equation}
    \gL = \sum_{\br \in \gR} \underbrace{\left\| \Hat{C}(\br) - C(\br) \right\|^2_2}_{\substack{\text{move output}\\\text{close to ground-}\\\text{truth image}}} + \overbrace{\lambda_{\rho_b}\gL_{\rho_b}}^{\substack{\text{move output}\\\text{close to correct}\\\text{border prob.}}} + \underbrace{\lambda_{\rho_s}\gL_{\rho_s}}_{\substack{\text{move output}\\\text{close to correct}\\\text{scattering dens.}}}
\label{eq:finalloss}
\end{equation}
where $\gR$ is the set of rays in a given batch, $\lambda_{\rho_b}$, $\lambda_{\rho_s}$ are the weighting factors, $C(\br)$ is the ground truth frame, $\Hat{C}(\br)$ is the frame obtained after rendering which we get by~\Cref{eq:unerfvol}. It is also straightforward to notice that incorporating our approach does not require any overhead during inference. Furthermore, once the diffusion model is trained, there is very little overhead (caused by inferencing the diffusion model) while training the NeRF.

\begin{table}[tb]
    \centering
    \begin{tabular}{lccc}
         \toprule
         \multicolumn{1}{c}{Method} & PSNR $\uparrow$ & SSIM $\uparrow$ & LPIPS $\downarrow$\\
         \midrule
         DCL-Net~\citep{guo2020sensorless} & 14.38 ($\pm$0.89) & 0.4418 ($\pm$0.16) & 0.5618 ($\pm$0.17) \\
         ImplicitVol~\citep{yeung2021implicitvol} & 18.42 ($\pm$0.83) & 0.6182 ($\pm$0.16) & 0.5482 ($\pm$0.06) \\
         Original NeRF~\citep{10.1145/3503250} & 14.90 ($\pm$1.06) & 0.5516 ($\pm$0.03) & 0.5371 ($\pm$0.04) \\
         \citet{gu2022representing} & 15.01 ($\pm$1.13) & 0.5724 ($\pm$0.18) & 0.5092 ($\pm$0.07) \\
         Instant-NGP~\citep{10.1145/3528223.3530127} & 22.73 ($\pm$4.06) & 0.7399 ($\pm$0.11) & 0.2842 ($\pm$0.08) \\
         TensoRF~\citep{10.1007/978-3-031-19824-3_20} & \cellcolor{tabsecond}25.14 ($\pm$6.16) & 0.7548 ($\pm$0.14) & 0.2716 ($\pm$0.11) \\
         Nerfacto~\citep{10.1145/3588432.3591516} & 15.04 ($\pm$1.76) & 0.4631 ($\pm$0.05) & 0.4645 ($\pm$0.07) \\
         Gaussian Splatting~\citep{kerbl20233d} & 22.33 ($\pm$4.59) & \cellcolor{tabthird}0.7746 ($\pm$0.13) & \cellcolor{tabthird}0.2683 ($\pm$0.10)\\
         Ultra-NeRF~\citep{pmlr-v227-wysocki24a} & \cellcolor{tabthird}24.14 ($\pm$2.54) & \cellcolor{tabsecond}0.8312 ($\pm$0.15) &\cellcolor{tabsecond}0.2573 ($\pm$0.16) \\
         \midrule
         Ours (\acronym) & \cellcolor{tabfirst}27.37 ($\pm$1.83) & \cellcolor{tabfirst}0.8412 ($\pm$0.13) & \cellcolor{tabfirst}0.2325 ($\pm$0.14)
         \\\bottomrule
    \end{tabular}
    \caption{\textbf{Quantitative Results.} We show quantitative comparisons between our \acronym\xspace against the baselines of our model on our ``Ultrasound in the wild'' dataset. We report the average PSNR $\uparrow$, SSIM $\uparrow$, LPIPS $\downarrow$ metrics across all scenes. The \colorbox{tabfirst}{best}, \colorbox{tabsecond}{second best}, and \colorbox{tabthird}{third best} results for each metric are color coded.}
    \label{tab:comparisions}
\end{table}

\section{Experiments}

\subsection{Ultrasound in the wild Dataset}
\label{sec:dataset}

We collect the data following a standardized protocol using a hand-held ultrasound device (Butterfly iQ+ by Butterfly Network Inc., Burlington, MA, USA\footnote{\url{https://www.butterflynetwork.com/iq-plus}}). We capture video and mid-sagittal images, including the suprapatellar longitudinal view of the suprapatellar recess of the knee. Using this, our pilot dataset captures $10$ unique casual sweeps at $30$ FPS on a subject around the human knee with at least $85$ frames in each sweep. All sweeps in this dataset have been captured by the authors on healthy knees following institutional REB (Research Ethics Board) guidelines; we provide details about REB guidelines in~\Cref{sec:ethics}. Following this, the camera registration is performed by COLMAP~\citep{10.1007/978-3-319-46487-9_31,Schonberger_2016_CVPR}. To create testing frames, we use every $8$th frame. A popular way to evaluate such models is to use another camera across a different trajectory to create the test frames, often with two imaging equipment attached to each other. However the nature of how ultrasounds are captured in the wild make it quite difficult to build a setup that could allow us to collect this data.

\subsection{Experimental Setup}

We evaluate the performance of our approaches based on the novel view synthesis quality on the ``Ultrasound in the wild'' Dataset (human knee) we introduced in~\Cref{sec:dataset} and the ``phantom dataset'' (human spine)~\citep{pmlr-v227-wysocki24a}. The generated images are compared with the ground truth test view images to calculate a few commonly used quantitative metrics that have become the standard way to evaluate NeRFs: PSNR, MS-SSIM~\citep{wang2003multiscale}, and LPIPS~\citep{Zhang_2018_CVPR}. We report the average values of these quantitative metrics across multiple frames, we follow the evaluation from most previous techniques~\citep{10.1145/3503250}. We compare our model with the baseline models of ImplicitVol~\citep{yeung2021implicitvol}, \citet{gu2022representing}, DCL-Net~\citep{guo2020sensorless}, Ultra-NeRF~\citep{pmlr-v227-wysocki24a}, as well as with standard NeRF methods: Original NeRF~\citep{10.1145/3503250}, Instant-NGP~\citep{10.1145/3528223.3530127}, TensoRF~\citep{10.1007/978-3-031-19824-3_20} and Nerfacto~\citep{10.1145/3588432.3591516}. We also compare our method with standard Gaussian Splatting~\citep{kerbl20233d}. We provide more details about the implementation in~\Cref{sec:implementation}.

\subsection{Results and Discussion}

\paragraph{Quantitative Results.}

\setlength{\tabcolsep}{1pt}
\begin{figure}[tb]
    \centering
    \begin{tabular}{ccccccc}
         && Nerfacto & G. Splatting & Ultra-NeRF & Ours (NeRF-US) & Ground Truth\\
        \midrule
         \multirow{1}{*}[16ex]{\rotatebox[origin=c]{90}{Scene 1}} &&  \includegraphics[width=0.192\textwidth]{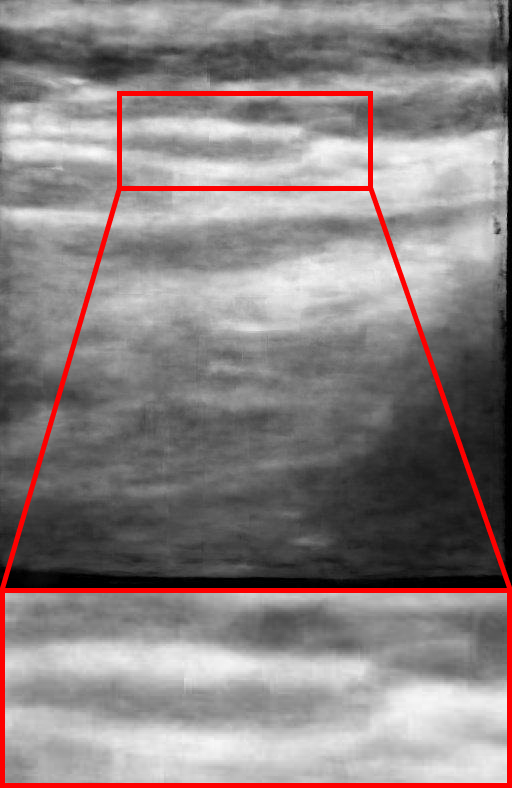} & \includegraphics[width=0.192\textwidth]{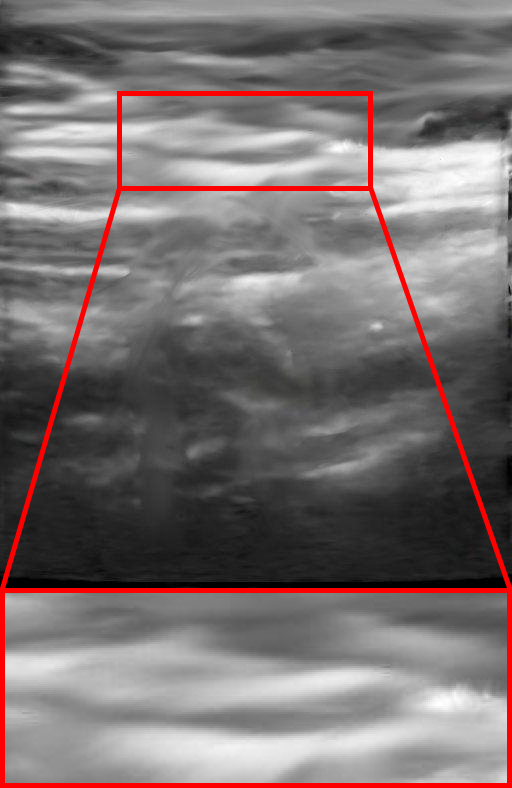} & \includegraphics[width=0.192\textwidth]{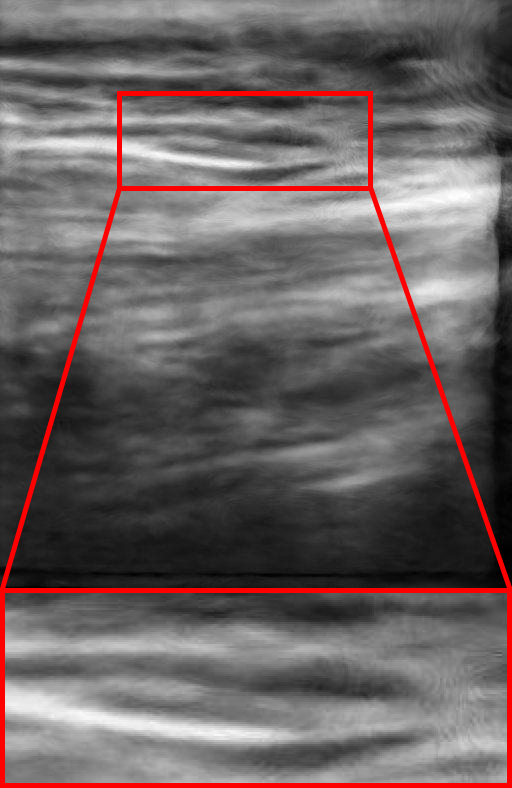} & \includegraphics[width=0.192\textwidth]{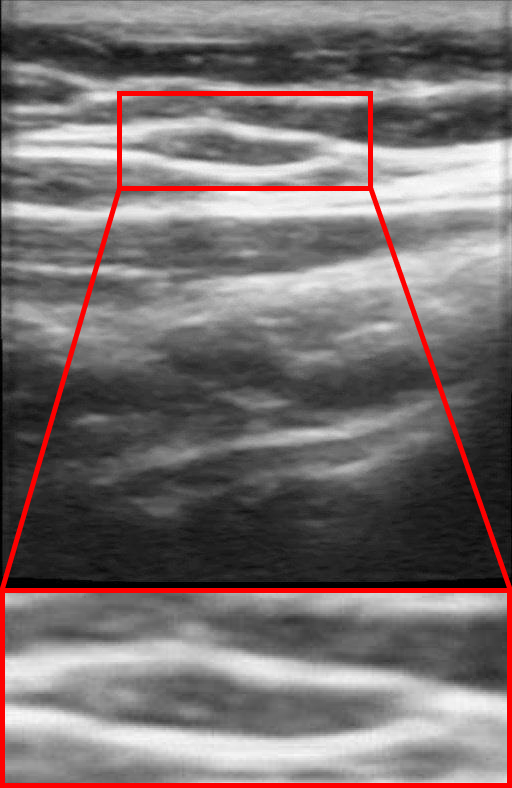} & \includegraphics[width=0.192\textwidth]{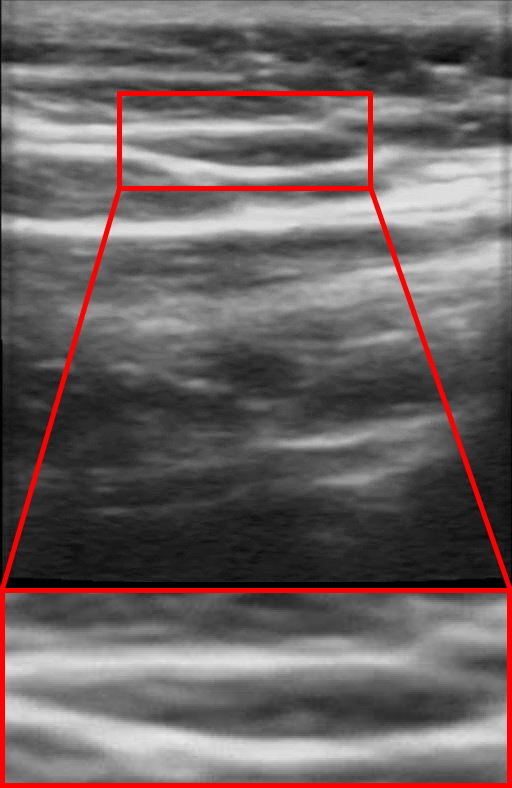} \\
         \multirow{1}{*}[16ex]{\rotatebox[origin=c]{90}{Scene 2}} && \includegraphics[width=0.192\textwidth]{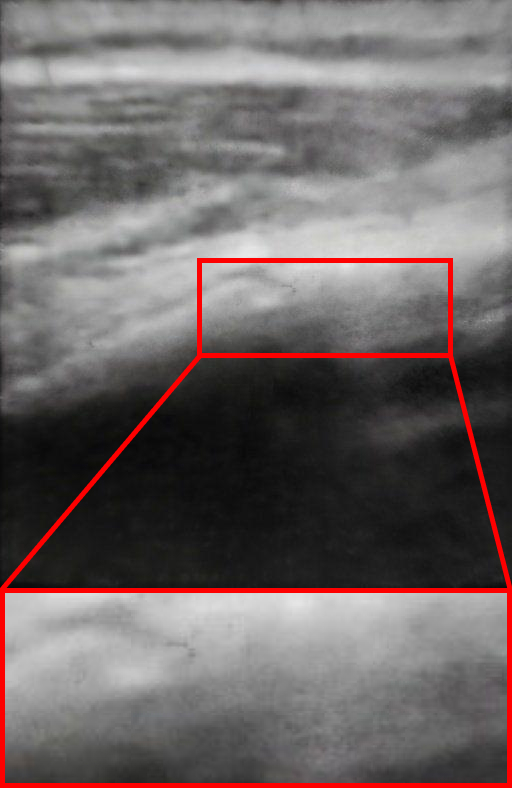} & \includegraphics[width=0.192\textwidth]{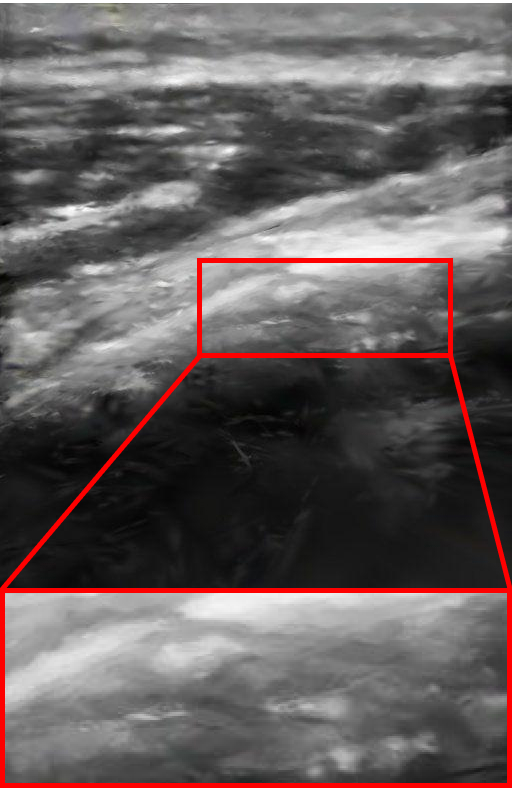} & \includegraphics[width=0.192\textwidth]{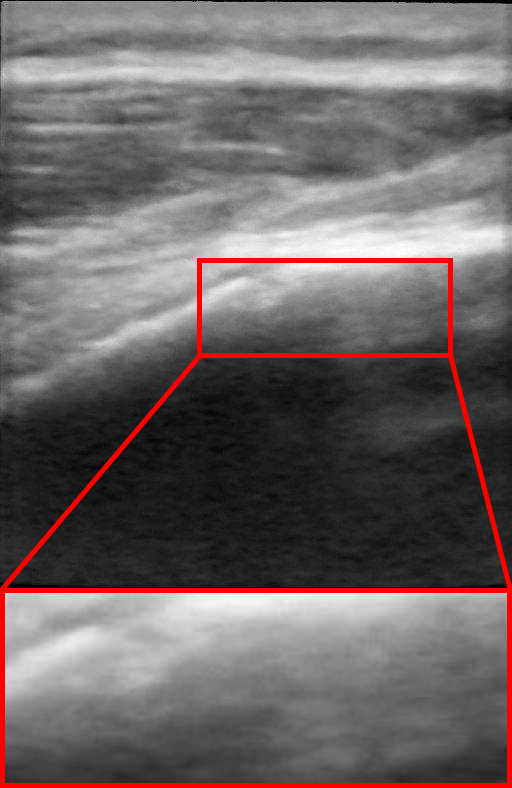} & \includegraphics[width=0.192\textwidth]{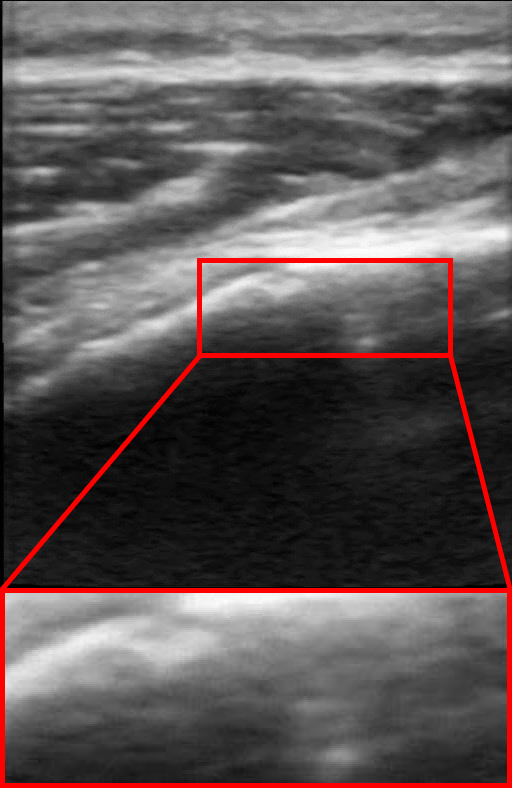} & \includegraphics[width=0.192\textwidth]{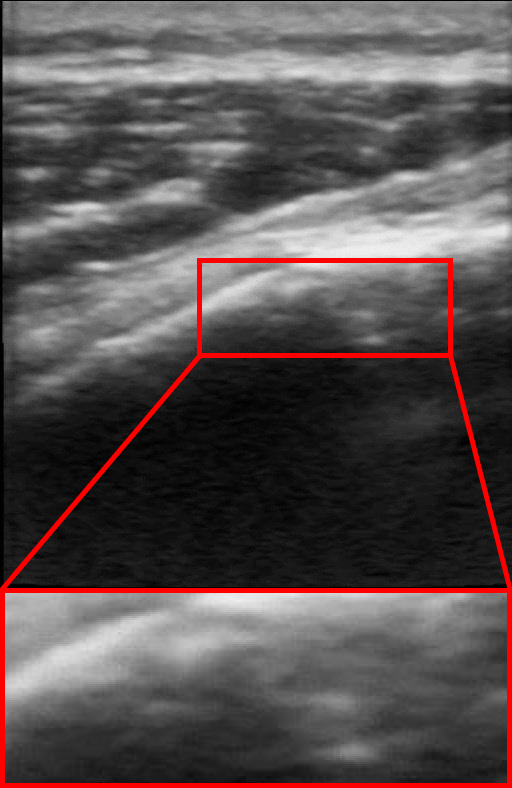} \\
         \multirow{1}{*}[16ex]{\rotatebox[origin=c]{90}{Scene 3}} && \includegraphics[width=0.192\textwidth]{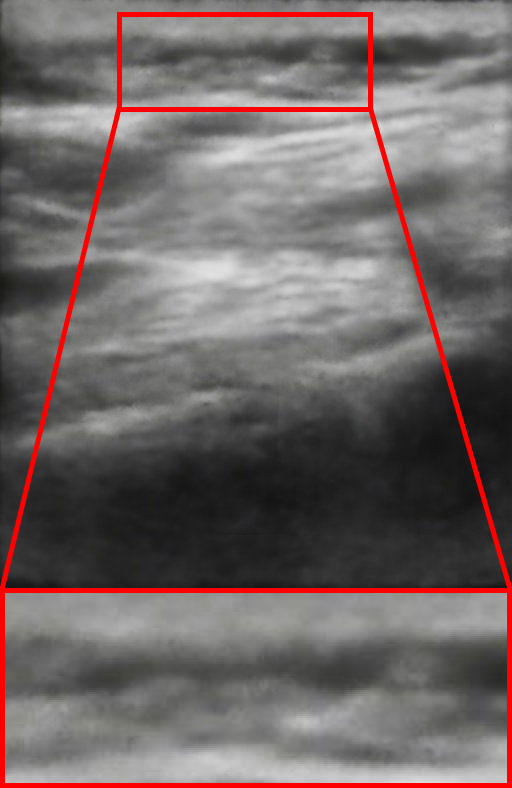} & \includegraphics[width=0.192\textwidth]{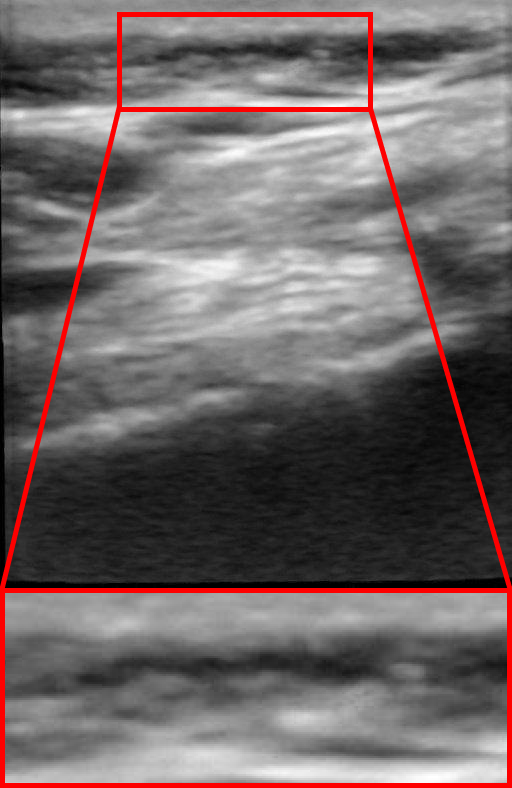} & \includegraphics[width=0.192\textwidth]{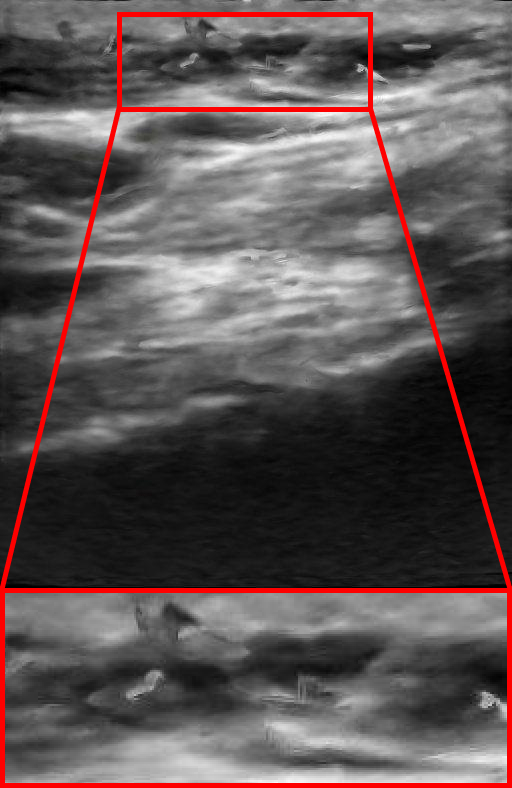} & \includegraphics[width=0.192\textwidth]{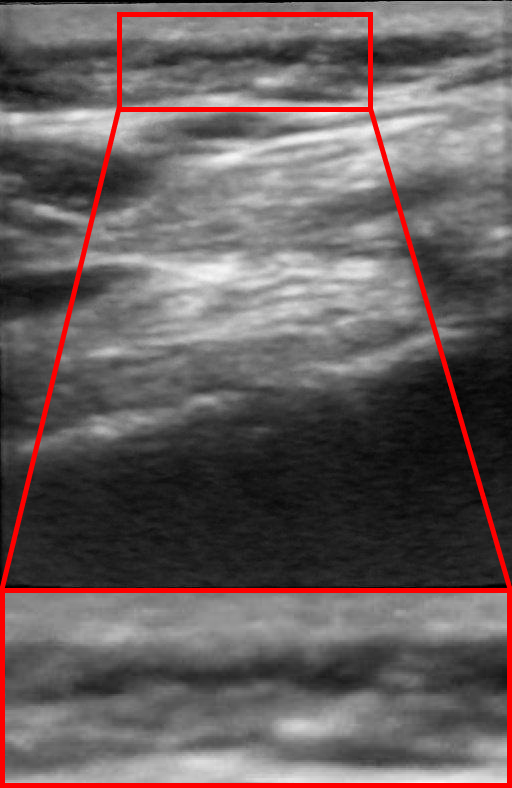} & \includegraphics[width=0.192\textwidth]{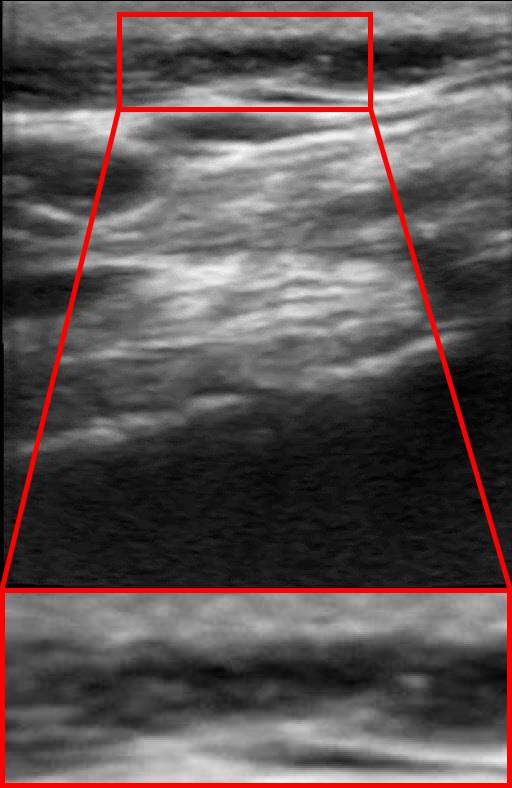}
    \end{tabular}
    \caption{\textbf{Qualitative Results.} We demonstrate the results of our method and compare it qualitatively with Nerfacto~\citep{10.1145/3588432.3591516}, Gaussian Splatting~\citep{kerbl20233d}, and Ultra-NeRF~\citep{pmlr-v227-wysocki24a}. Our approach, NeRF-US, produces accurate and high-quality reconstructions as compared to the baseline models on novel views (\textbf{best viewed with zoom}).}
    \label{fig:results}
\end{figure}

\begin{table}[tb]
    \centering
    \begin{tabular}{lccc}
         \toprule
         \multicolumn{1}{c}{Method} & PSNR $\uparrow$ & SSIM $\uparrow$ & LPIPS $\downarrow$\\
         \midrule
         Original NeRF~\citep{10.1145/3503250} & 21.36 ($\pm$1.27) & 0.5100 ($\pm$0.12) & 0.2415 ($\pm$0.14) \\
         Instant-NGP~\citep{10.1145/3528223.3530127} & 26.13 ($\pm$1.35) & 0.5052 ($\pm$0.05) & 0.2324 ($\pm$0.06) \\
         TensoRF~\citep{10.1007/978-3-031-19824-3_20} & 27.16 ($\pm$3.17) & 0.5178 ($\pm$0.03) & 0.2342 ($\pm$0.05) \\
         Nerfacto~\citep{10.1145/3588432.3591516} & 26.28 ($\pm$2.63) & 0.5124 ($\pm$0.04) & 0.2353 ($\pm$0.04) \\
         Gaussian Splatting~\citep{kerbl20233d} & \cellcolor{tabfirst}29.72 ($\pm$4.27) & \cellcolor{tabthird}0.5248 ($\pm$0.12) & \cellcolor{tabthird}0.2214 ($\pm$0.11) \\
         Ultra-NeRF~\citep{pmlr-v227-wysocki24a} & \cellcolor{tabthird}28.12 ($\pm$2.35) & \cellcolor{tabfirst}0.5314 ($\pm$0.09) & \cellcolor{tabsecond}0.2182 ($\pm$0.10) \\
         \midrule
         Ours (\acronym) & \cellcolor{tabsecond}29.24 ($\pm$2.13) & \cellcolor{tabsecond}0.5306 ($\pm$0.05) & \cellcolor{tabfirst}0.1963 ($\pm$0.13)
         \\\bottomrule
    \end{tabular}
    \caption{\textbf{Quantitative Results.} We show quantitative comparisons between our \acronym\xspace against the baselines of our model on the ``phantom'' dataset from~\citet{pmlr-v227-wysocki24a}. However, the captures in this dataset are not in the wild. We report the average PSNR $\uparrow$, SSIM $\uparrow$, LPIPS $\downarrow$ metrics across all scenes. The \colorbox{tabfirst}{best}, \colorbox{tabsecond}{second best}, and \colorbox{tabthird}{third best} results for each metric are color coded.}
    \label{tab:comparisions-undata}
\end{table}

We demonstrate the quantitative results of our approach against other methods for novel views on our ``Ultrasound in the wild'' dataset. We compare our approach quantitatively across methods that are specialized for 3D ultrasound reconstruction as well as state-of-the-art standard 3D reconstruction methods which are not specific to ultrasound or medical imaging. We particularly compare with state-of-the-art 3D reconstruction methods which are not focused on ultrasound to demonstrate our approach's significance for reconstruction such sound-based imaging. We show these quantitative results and comparisions in~\Cref{tab:comparisions} and~\Cref{tab:comparisions-undata}. We also demonstrate the quantitative results on the ``phantom'' dataset~\citep{pmlr-v227-wysocki24a}. However, this dataset does not feature in the wild captures.

\begin{figure}[tb]
    \centering
    \begin{tabular}{ccccccc}
         && Nerfacto & G. Splatting & Ultra-NeRF & Ours (NeRF-US) & Ground Truth\\
        \midrule
         \multirow{1}{*}[12.5ex]{\rotatebox[origin=c]{90}{Scene 2}} && \includegraphics[width=0.192\textwidth]{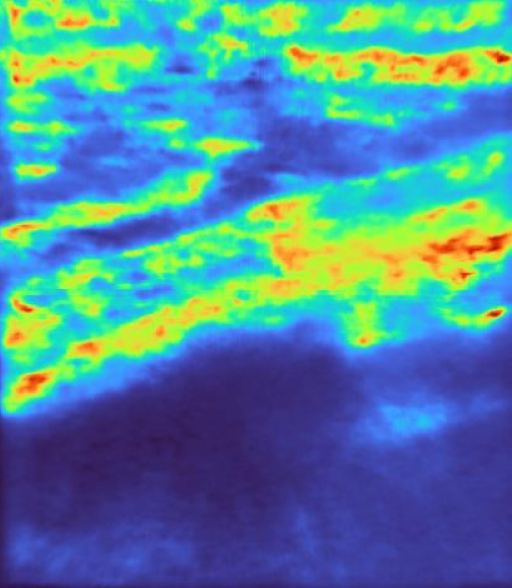} & \includegraphics[width=0.192\textwidth]{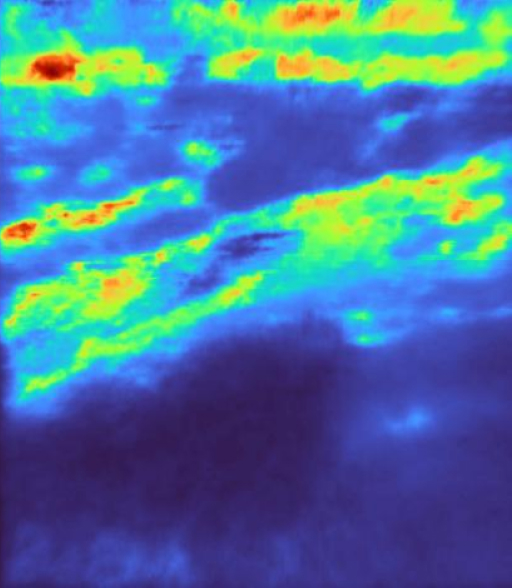} & \includegraphics[width=0.192\textwidth]{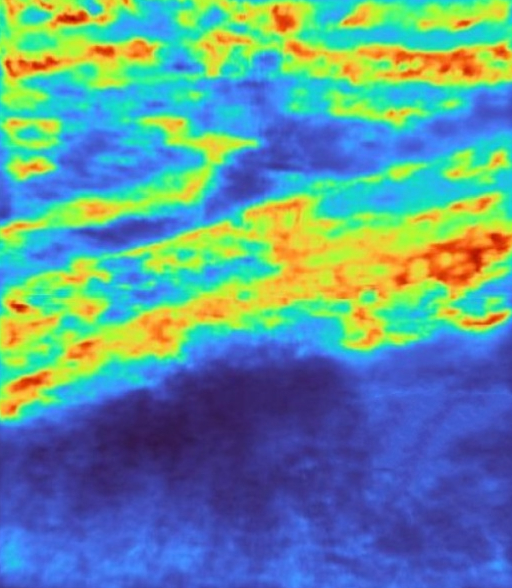} & \includegraphics[width=0.192\textwidth]{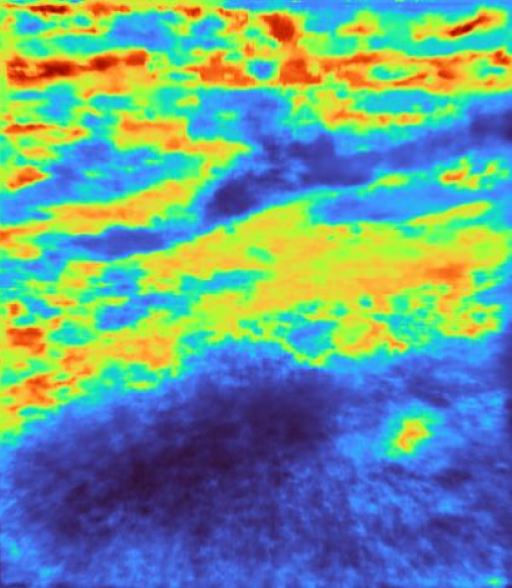} & \includegraphics[width=0.192\textwidth]{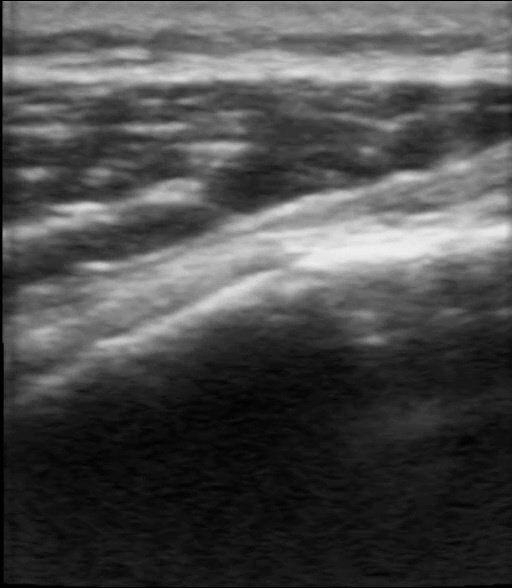} \\
         \multirow{1}{*}[12.5ex]{\rotatebox[origin=c]{90}{Scene 3}} && \includegraphics[width=0.192\textwidth]{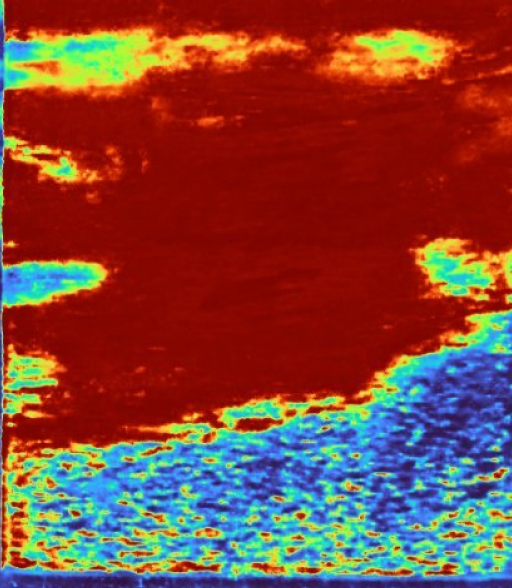} & \includegraphics[width=0.192\textwidth]{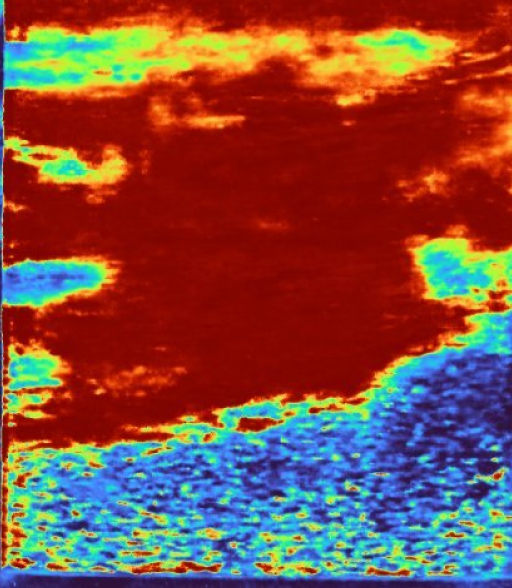} & \includegraphics[width=0.192\textwidth]{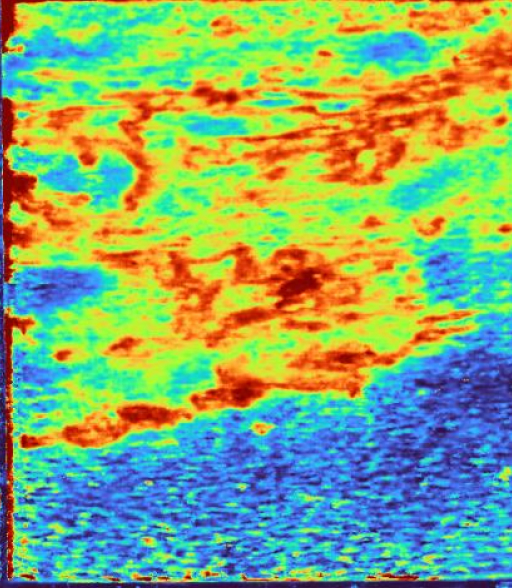} & \includegraphics[width=0.192\textwidth]{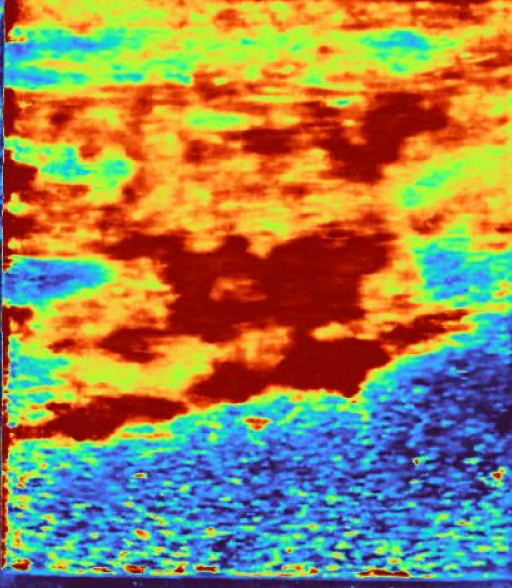} & \includegraphics[width=0.192\textwidth]{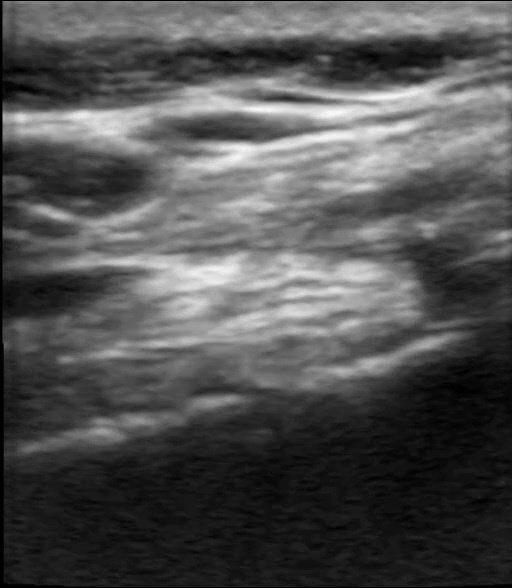}
    \end{tabular}
    \caption{\textbf{Qualitative Results.} We demonstrate the results of depth maps produced from our method and compare them qualitatively with Nerfacto~\citep{10.1145/3588432.3591516}, Gaussian Splatting~\citep{kerbl20233d}, and Ultra-NeRF~\citep{pmlr-v227-wysocki24a} (\textbf{best viewed in color and
with zoom}).}
    \label{fig:boundaries}
\end{figure}

\paragraph{Qualitative Results.} We present the qualitative results of
novel view synthesis in~\Cref{fig:results}. We particularly show that other approaches tend to reconstruct ultrasound scenes with severe geometric artifacts. The rendered ultrasound scenes are blurry or torn apart along the moving trajectory. We particularly also notice that other approaches especially Ultra-NeRF~\citep{pmlr-v227-wysocki24a} produce reconstructions where tissue borders and separation are not captured properly due to a lot of uncertainty in these areas by all previous approaches. We also demonstrate some of these issues in~\Cref{fig:boundaries}. Our approach particularly alleviates these issues with the reconstruction of 3D ultrasound representations.

\setlength{\tabcolsep}{1pt}

\subsection{Ablations}

\begin{figure}[t]
    \centering
    \begin{tabular}{ccccc}
        w/o $\gL_{\rho_b}$ & w/o $\gL_{\rho_s}$ & w/o $I(t)$ & \acronym & Ground Truth\\
        \midrule
        \includegraphics[width=0.195\textwidth]{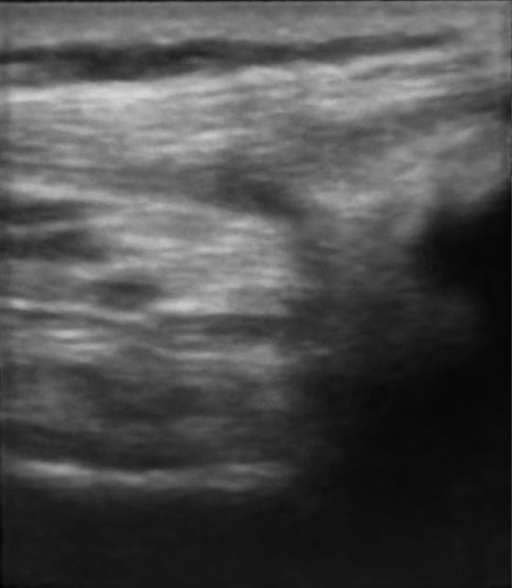} & \includegraphics[width=0.195\textwidth]{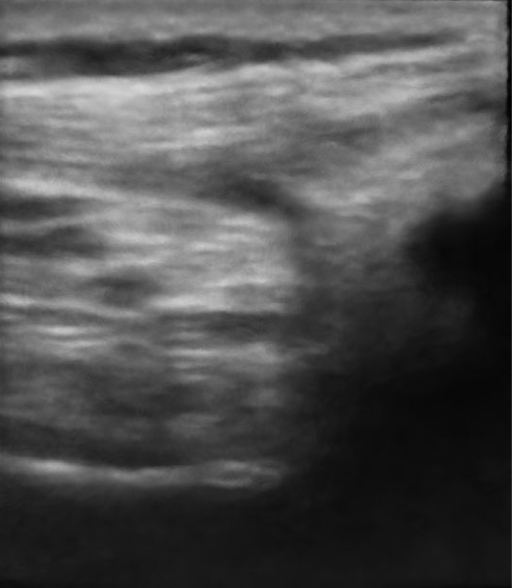} & \includegraphics[width=0.195\textwidth]{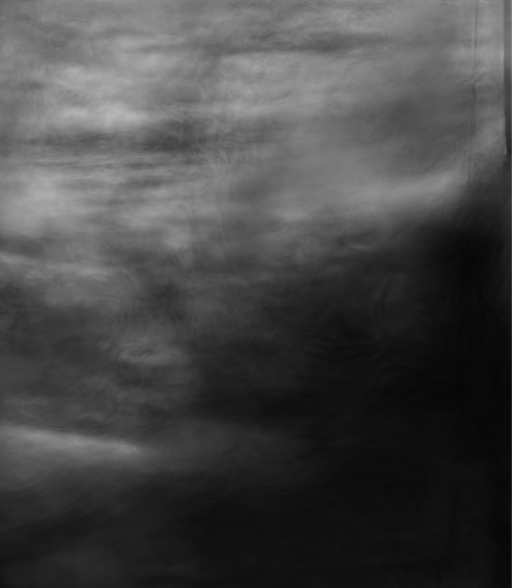} & \includegraphics[width=0.195\textwidth]{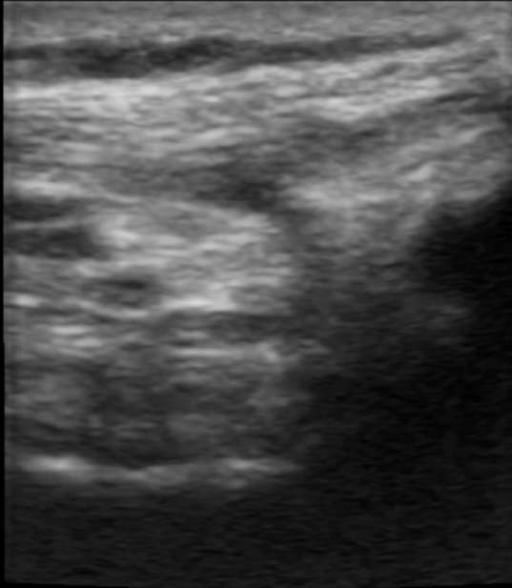} & \includegraphics[width=0.195\textwidth]{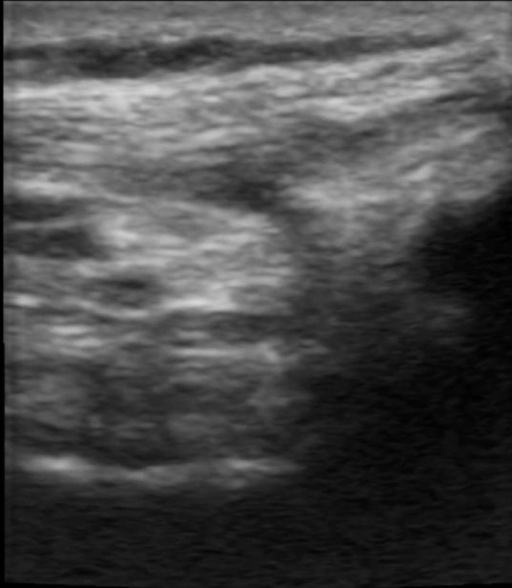}
    \end{tabular}
    \caption{\textbf{Ablation Study.} We assess the qualitative impact of removing each of the components of our pipeline: $\gL_{\rho_b}$, $\gL_{\rho_s}$, and $I(t)$. Our approach without the border probability guidance ($\gL_{\rho_b}$) fail to reconstruct minor details in the scene especially it is unable to clearly segregate multiple objects in the image. Removing the scattering density guidance ($\gL_{\rho_s}$) produces very blurred reconstructions. Our approach without the ultrasound rendering produces a lot of artifacts in the reconstruction and fails to capture most aspects of the reconstruction. (\textbf{Best viewed with zoom}).}
    \label{fig:ablations}
\end{figure}

We provide an in-depth analysis motivating our two-fold training approach highlighted in~\Cref{sec:methods} by ablating each component and evaluating the use of each component. We provide ablations in~\Cref{fig:ablations} and~\Cref{tab:ablations}.

\paragraph{Border Probability Guidance (w/o $\gL_{\rho_b}$).} To implement this we simply set $\lambda_{\rho_b}$ to $0$ for the entire training process in~\Cref{eq:finalloss}. We find that the border probability guidance is useful for accurately modeling tissue interface locations and border locations; if we disable it during training, we observe that many objects blend into each other and have discontinuities in the representation space.

\paragraph{Scattering Density Guidance (w/o $\gL_{\rho_s}$).} To implement this we simply set $\lambda_{\rho_s}$ to $0$ for the entire training process in~\Cref{eq:finalloss}. We find that the scattering density guidance is useful to accurately model bodily objects; if we disable it during the training, we observe that many microstructural features are missing from the reconstructions. For example, in~\Cref{fig:ablations}, many minute details are entirely missing when we do not use scattering density guidance.

\paragraph{Ultrasound Rendering (w/o $I(t)$).} Technically, learning a NeRF model without the ultrasound rendering (\Cref{eq:unerfvol}) is possible, to implement this we still calculate the rendered 2D image with ultrasound rendering since the results from this are used to calculate $\gL_{\rho_b}$ and $\gL_{\rho_s}$ terms, however, we parallelly also learn another model which is trained with the objective function as~\Cref{eq:finalloss} but uses standard volumetric rendering to calculate $\hat{C}{r}$ and we report the metrics for this NeRF model. We find that using ultrasound rendering is useful to model ultrasound effects on the image; if we disable it during the training, we observe that while individual frames look decently reconstructed, the 3D geometry of the underlying objects is improper due to the nature of how ultrasounds capture data.

\begin{table}[tb]
    \centering
    \setlength{\tabcolsep}{5pt} 
    \begin{tabular}{lccc}
         \toprule
         \multicolumn{1}{c}{Method} & PSNR $\uparrow$ & SSIM $\uparrow$ & LPIPS $\downarrow$\\
         \midrule
         Ultra-NeRF~\citep{pmlr-v227-wysocki24a}\xspace\xspace& 24.14 ($\pm$2.54) & \cellcolor{tabsecond}0.8312 ($\pm$0.15) & \cellcolor{tabsecond}0.2573 ($\pm$0.16) \\
         Ours (\acronym) & \cellcolor{tabfirst}27.37 ($\pm$1.83) & \cellcolor{tabfirst}0.8412 ($\pm$0.13) & \cellcolor{tabfirst}0.2325 ($\pm$0.14) \\
         \midrule
         \emph{Ablation Study}&&&\\
         \midrule
         Ours (\acronym) & & & \\
         \quad w/o $\gL_{\rho_b}$ & \cellcolor{tabthird}24.89 ($\pm$1.72) & \cellcolor{tabthird}0.8208 ($\pm$0.98) & 0.2613 ($\pm$0.13) \\
         \quad w/o $\gL_{\rho_s}$ & \cellcolor{tabsecond}26.02 ($\pm$1.67) & 0.8197 ($\pm$1.10) & 0.2598 ($\pm$0.08) \\         
         \quad w/o $I(t)$ & 23.98 ($\pm$1.46) & 0.8065 ($\pm$1.23) & \cellcolor{tabthird}0.2591 ($\pm$0.11)
         \\\bottomrule
    \end{tabular}
    \caption{\textbf{Ablation Study.} We show ablations of our approach: w/o $\gL_{\rho_b}$, w/o $\gL_{\rho_s}$, and w/o the rendering $I(t)$. The \colorbox{tabfirst}{best}, \colorbox{tabsecond}{second best}, and \colorbox{tabthird}{third best} results for each metric are color coded.}
    \label{tab:ablations}
\end{table}

\setlength{\tabcolsep}{1pt}

\section{Limitations}

Although our method produces compelling
reconstructions of ultrasound imaging, there are several limitations and avenues for future work. First, the complexity of motion in our scenes is limited to simple movements, while our approach is accurately able to reconstruct ultrasounds taken in the natural setting which always has some movement, our method fails to reconstruct accurately when there are immensely complex full-body motions. We believe that our method will directly benefit from the progress on dynamic reconstruction methods that use dynamic representations eg. HyperNeRF~\citep{park2021hypernerf}, NeRF-DS~\citep{10204790}, Dynamic 3D Gaussians~\citep{luiten2023dynamic}. Moreover, currently our diffusion model is fine-tuned on voxels specific to a body region, while fine-tuning the diffusion model is very cheap ($<5$ hours on A$100\text{-}80$), and we believe that an interesting avenue for future work would be to build a generalized diffusion model. Lastly, our diffusion model is fine-tuned on a dataset of synthetic voxels (notice that our NeRF model is evaluated and trained on real-life data only), and we currently do not experiment with a real-life dataset from 3D captures. Furthermore, an interesting avenue for future works is to explore the effect of training the diffusion model with other modalities, we particularly believe it might be promising to explore training the diffusion model on Magnetic Resonance Imaging (MRI) scans to reconstruct ultrasound imaging.

\section{Conclusion}

Our work builds a NeRF-based technique that can perform accurate view synthesis and 3D reconstruction on ultrasound imaging. Our work is the first to tackle the problem of performing view synthesis and 3D reconstructions on ultrasound imaging data collected in its natural in the wild form as opposed to works that tackle this problem on simulated data or very heavy ultrasound capture mechanisms. We also observe accurate artifact-free outputs for ultrasound imaging with our method. We will release the code and data to facilitate future research in this area. 

\ifarxivorcameraready
\newpage
    \acks{This research was enabled in part by support provided by the Digital Research Alliance of Canada \footnote{\url{https://alliancecan.ca/}}. This research was supported in part with Cloud TPUs from Google's TPU Research Cloud (TRC)\footnote{\url{https://sites.research.google/trc/about/}}. The resources used to prepare this research were provided, in part, by the Province of Ontario, the Government of Canada through CIFAR, and companies sponsoring the Vector Institute \footnote{\url{https://vectorinstitute.ai/partnerships/current-partners/}}. We thank~\citet{pmlr-v227-wysocki24a} for readily sharing the data they collected for their work. We thank anonymous reviewers of the MLHC Conference for their insightful suggestions which we incorporate in this work.}
\fi

\bibliography{main}

\newpage
\appendix

\section{Implementation Details}
\label{sec:implementation}

\subsection{Training the Diffusion Model}

The pre-trained diffusion model is a UNet-based diffusion model designed for $32\times32\times32$ data cubes. The model utilizes 3D convolutions with a base channel count of 32, employing channel multipliers of (1, 2, 4, 8) across resolution levels. We use two residual blocks per resolution level, with self-attention mechanisms strategically applied at 16$\times$ and 8$\times$ downsampling. The model also incorporates time embeddings. The model features a downsampling path that progressively reduces spatial dimensions while increasing channel count, a bottleneck with residual connections and attention mechanisms, and an upsampling path with skip connections. The pre-trained model also uses scale conditioning through a learned embedding.

We provide more details on how we perform LoRA~\citep{hu2022lora} adaption to the denoising diffusion model. Our implementation closely follows the original LoRA, which demonstrated the adaption approach for language models, and draws inspiration from the HuggingFace Diffusers implementation, which demonstrates adaption for text-to-image (T2I) models.

We use the 3D diffusion model $\Phi$ with parameters $\theta$ trained on ShapeNet~\citep{chang2015shapenet} which works on $32\times 32\times 32$ occupancy grid, $x$, from Nerfbusters~\citep{Nerfbusters2023}. Based on LoRA~\citep{hu2022lora}, for a given layer in $\Phi$, we introduce trainable low-rank matrices. For the weight matrix $W \in \mathbb{R}^{d_{\text{out}} \times d_{\text{in}}}$ in the model, LoRA decomposes the adaptation into two low-rank matrices, $A \in \mathbb{R}^{d_{\text{out}} \times r}$ and $B \in \mathbb{R}^{r \times d_{\text{in}}}$, where $r \ll \min(d_{\text{out}}, d_{\text{in}})$ representing the rank controlling adaption. The updated weights $W'$ can be computed as,
\begin{equation}
    W' = W + \delta(AB)
\label{eq:wupdate}
\end{equation}

where $\delta$ is a scaling factor that determines the magnitude of the adaptation and $AB$ represents a low-rank update to the original weights. 

Our fine-tuning process now utilizes the same loss function as the original DDPM training, with the low-rank update applied to the model parameters, giving us~\Cref{eq:ft}.

\subsection{NeRF Optimization Objective}

Our optimization objective for the NeRF was defined in~\Cref{eq:finalloss}, we can write this optimization objective as,
\begin{equation}
\begin{split}
    \gL &= \sum_{\br \in \gR} \overbrace{\left\| \Hat{C}(\br) - C(\br) \right\|^2_2}^{\text{photometric loss}} + \lambda_{\rho_b}\gL_{\rho_b} + \lambda_{\rho_s}\gL_{\rho_s}\\
    &= \sum_{\br \in \gR} \left\| \Hat{C}(\br) - C(\br) \right\|^2_2 + \lambda_{\rho_b}\left(\frac{1}{N}\sum_{i=1}^N \left\lvert C(i)^{\rho_b} - \Phi_{\theta+\delta(AB)}(C(i))^{\rho_b} \right\rvert^2\right) \\
    & \hspace{10em}+ \lambda_{\rho_s}\left(\frac{1}{N}\sum_{i=1}^N \left\lvert C(i)^{\rho_s} - \Phi_{\theta+\delta(AB)}(C(i))^{\rho_s} \right\rvert^2\right)
\end{split} 
\end{equation}

We can notice that while our loss formulation is different than the Density Score Distillation Sampling~\citep{Nerfbusters2023}, however it can be still looked upon as adding a regularizer to the original NeRF photometric loss that penalizes an incorrect border probability and scattering density. Unlike Ultra-NeRF~\citep{pmlr-v227-wysocki24a}, we do not use a SSIM loss, we observe that with the new definition of $\gL$, using the SSIM loss does not demonstrate any benefits over using the photometric loss term. Furthermore, we also observe that setting $\lambda_{\rho_b} > \lambda_{\rho_s}$ usually works well towards reconstructing high-quality scenes.

\subsection{NeRF MLP}

Following previous work on NeRFs like mip-NeRF~\citep{Barron_2022_CVPR}, we use an MLP $(F_\Theta)$ with $8$ layers and $256$ hidden layer units with ReLU and a skip connection between the input and the fifth layer. To ensure numerical stability and physiologically plausible results, we enforce the attenuation parameter to assume continuous positive values, while reflectance, border probability, scattering density, and scattering intensity are confined within the range of $[0, 1]$. We do not employ additional MLP directional components for camera viewing directions $(\theta,\phi)$. This decision stems from the recognition that our MLP, does not directly assimilate information from the camera viewing angles. Rather, this aspect is effectively addressed by the ultrasound rendering of the reconstruction pipeline. Following NeRF~\citep{10.1145/3503250}, we also use positional encodings to train the MLP.

\subsection{Training Hyperparameters}

All of our final code was optimized for a 1 x A100-80 GB GPU. For our Diffusion model, we trained our model for $30K$ steps with a batch size of $32$ on the $32\times 32\times 32$ resolution. We use the Adam optimizer~\citep{kingma2017adam} with $\beta_1=0.9$, $\beta_2=0.999$, and $\epsilon=10^{-8}$ with an initial learning rate of $10^{-4}$. We use cosine decay for the learning rate. We use a rank of $4$ for the LoRA update weights. Our NeRF implementation is based upon Nerfstudio~\citep{10.1145/3588432.3591516}. We trained our model for $300K$ iterations with a batch size of $2^{12}$. We use the RAdam optimizer~\citep{Liu2020On} with $\beta_1=0.9$ and $\beta_2=0.999$. We use a learning rate that starts at $5\times10^{-4}$ and decays exponentially to $5\times10^{-5}$, and $\epsilon=10^{-8}$. We however, notice that the results become pretty accurate after the first $100K$ iterations.

\subsection{Code and Data}

\ifdefined\submission
    To foster future work in this direction, we commit to open-sourcing our code and ultrasound in the wild dataset upon acceptance. We do not share these here to comply with the double-blind review.
\fi
\ifarxivorcameraready

    To foster future work in this direction we open-source our code and Ultrasound in the wild dataset which can be found on our project page at: \href{https://\website}{\url{\website}}.
\fi

\subsection{Baseline Models}

Here we share the details for the baseline models we compare our approach with in~\Cref{tab:comparisions}. Our baseline models for Original NeRF~\citep{10.1145/3503250}, Instant-NGP~\citep{10.1145/3528223.3530127}, TensoRF~\citep{10.1007/978-3-031-19824-3_20}, Nerfacto~\citep{10.1145/3588432.3591516}, and Gaussian Splatting~\citep{kerbl20233d} are trained with Nerfstudio~\citep{10.1145/3588432.3591516} implementations using the details we list below. Just like rest of our code all of our baseline models were trained on 1 x A100-80 GB GPU.

\paragraph{DCL-Net.} We trained this baseline model for $300$ epochs with a batch size of $2^{5}$ using the Adam optimizer~\citep{kingma2017adam} with $\beta_1=0.9$, $\beta_2=0.999$, and $\epsilon=10^{-14}$. We use cosine decay starting with the learning rate of $5\times10^{-5}$ and a warmup of $5$ epochs. Furthermore, all the underlying images are resized to $224 \times 224$-sized image.

\paragraph{ImplicitVol.} We construct this baseline model with an MLP of $5$ layers with $128$ neurons. We trained this baseline model for $10K$ epochs and also use NeRF-style positional encodings. Since using this technique involves first estimating 2D plane locations, as suggested by ImplicitVol in their work we first estimate these plane positions using PlaneInVol~\citep{wang2021nerf}. We use the Adam optimizer~\citep{kingma2017adam} with $\beta_1=0.9$, $\beta_2=0.999$, and $\epsilon=10^{-8}$ with an initial learning rate of $10^{-3}$. We use the standard multi-step schedule with each of its stages at every $10$ epochs with $\gamma=0.9954$. We incorporate a window size of $k=5$ for calculating the SSIM loss.

\paragraph{Original NeRF.} We trained this baseline model for $300K$ iterations with a batch size of $2^{12}$. We use the RAdam optimizer~\citep{Liu2020On} with $\beta_1=0.9$ and $\beta_2=0.999$. We use a learning rate that starts at $5\times10^{-4}$ and decays exponentially to $5\times10^{-5}$, and $\epsilon=10^{-8}$. We also use gradient scaling to ensure that gradients near the camera are scaled down. We also use hierarchical sampling with 64 coarse samples and 128 importance samples for fine field evaluation. We use the NeRF defaults for all other hyperparamters.

\paragraph{\citet{gu2022representing}.} We trained this baseline model for $20K$ iterations with a batch size of $2^{2}$. We use the Adam optimizer~\citep{kingma2017adam} with an initial learning rate of $10^{-4}$, $\epsilon=10^{-8}$, $\beta_1=0.9, \beta_2=0.999$ and a meta learning rate of $10^{-5}$. We also use LookAhead with $10$ steps. We use an $\omega=240$ for the SIREN activation. We use the defaults from \citet{gu2022representing} for all other hyperparamters.

\paragraph{Instant-NGP.} We trained this baseline model for $30K$ iterations with a batch size of $2^{12}$. We use the Adam optimizer~\citep{kingma2017adam}  with $\beta_1=0.9$ and $\beta_2=0.999$. We use an initial learning rate of $10^{-2}$ and $\epsilon=10^{-15}$. We exponentially decay the learning rate from $10^{-2}$ to $10^{-4}$ and use no warmup. We use a grid resolution of $128$ with $4$ grid levels for the multiresolution hash encoding. We also use a resolution of $2^{11}$ for the hashmap used by the MLP. We perform the sampling from the rays in between $[5\times10^{-2},10^3]$. We use the Instant-NGP defaults for all other hyperparamters.

\paragraph{TensoRF.} We trained this baseline model for $30K$ iterations with a batch size of $2^{12}$. We use the Adam optimizer~\citep{kingma2017adam}  with $\beta_1=0.9$ and $\beta_2=0.999$. We use an initial learning rate of $10^{-2}$ and $\epsilon=10^{-8}$. We exponentially decay the learning rate from $10^{-3}$ to $10^{-4}$ and use no warmup. We also use TV regularization while training the field. We use the TensoRF defaults for all other hyperparamters.

\paragraph{Nerfacto.} We trained this baseline model for $30K$ iterations with a batch size of $2^{12}$. We use the Adam optimizer~\citep{kingma2017adam} with $\beta_1=0.9$ and $\beta_2=0.999$. We use an initial learning rate of $10^{-2}$ and $\epsilon=10^{-15}$. We exponentially decay the learning rate from $10^{-2}$ to $10^{-4}$ and use no warmup. We use a vector with 6 parameters representing SO3$\times$R3 map for the camera optimizer. We use the Nerfacto defaults for all other hyperparamters.

\paragraph{Gaussian Splatting.} We trained this baseline model for $30K$ iterations with a batch size of $2^{12}$. We use the Adam optimizer~\citep{kingma2017adam} with $\beta_1=0.9$ and $\beta_2=0.999$ for the Gaussian means network. For the Gaussian means network exponentially decay the learning rate from $1.6\times10^{-4}$ to $1.6\times10^{-6}$, and $\eps=10^{-15}$. We set the threshold for frustum culling Gaussians to $5\times10^{-3}$. For rendering we use the EWA volume splatting with a $[0.3, 0.3]$ screen space blurring kernel. Following recent popular work with Gaussian Splatting, we also apply a regularization loss when the ratio of a Gaussian's maximum to minimum scale exceeds a threshold. We use the Gaussian Splatting defaults for all other hyperparamters.

\paragraph{Ultra-NeRF.} We trained this baseline model for $100K$ iterations with a batch size of $2^{12}$. We use the Adam optimizer~\citep{Liu2020On} with $\beta_1=0.9$ and $\beta_2=0.999$. We use a learning rate that starts at $10^{-4}$ and decays exponentially in $250K$ steps with a decay of $0.1$, and $\epsilon=10^{-8}$. For the loss calculation we use a window size of $k=7$ and $\lambda=0.7$. We use the NeRF defaults for all other hyperparamters.

\section{Evaluation Metrics}

To evaluate the quality of our rendered images, we employ three standard metrics commonly used in the assessment of such models.

\paragraph{PSNR} quantifies the ratio between the maximum possible signal power and the power of distorting noise, with higher values indicating better reconstruction quality.

\paragraph{SSIM} evaluates the perceived quality of images by considering structural information changes, with values closer to 1 indicating higher similarity.

\paragraph{LPIPS} is a perceptual metric leveraging deep neural networks trained on human judgments, aims to capture perceptual similarities between image patches, with lower scores indicating greater perceptual similarity.

\section{Additional Qualitative Results}

\begin{figure}[htb]
    \centering
    \begin{tabular}{cccc}
    \includegraphics[width=0.25\textwidth]{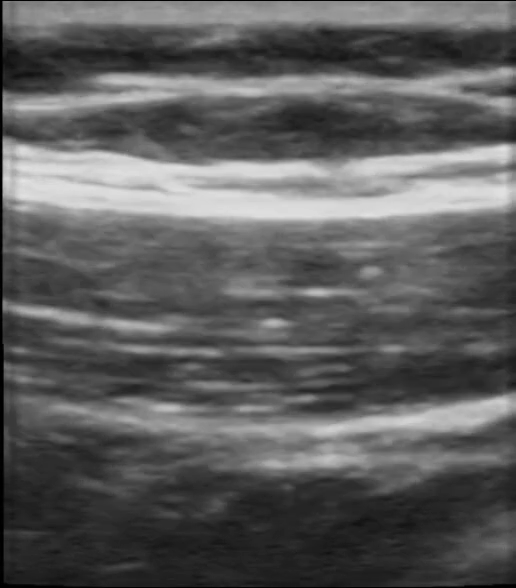} & \includegraphics[width=0.25\textwidth]{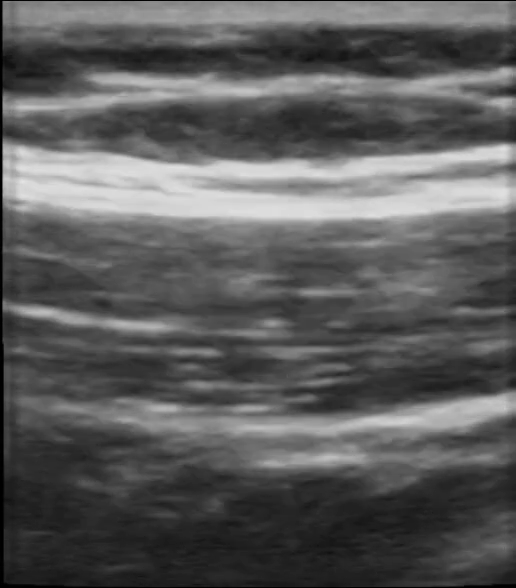} & \includegraphics[width=0.25\textwidth]{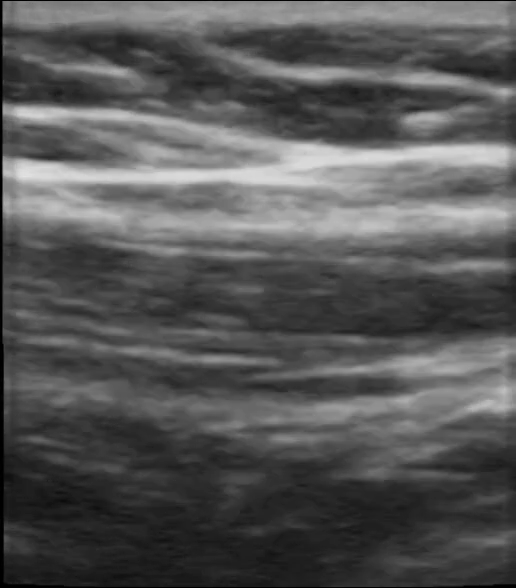} &
    \includegraphics[width=0.25\textwidth]{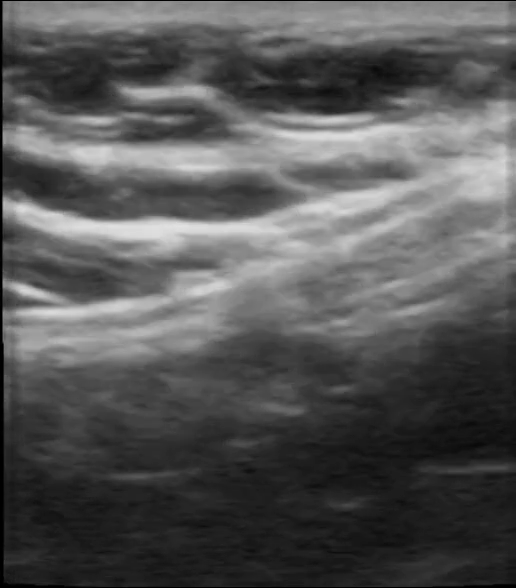} \\ \includegraphics[width=0.25\textwidth]{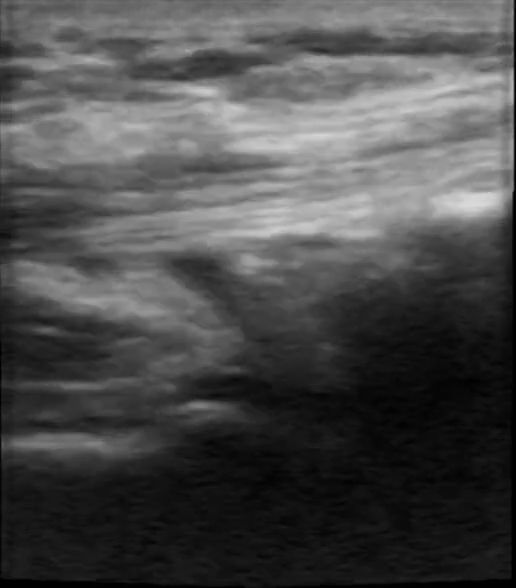} & \includegraphics[width=0.25\textwidth]{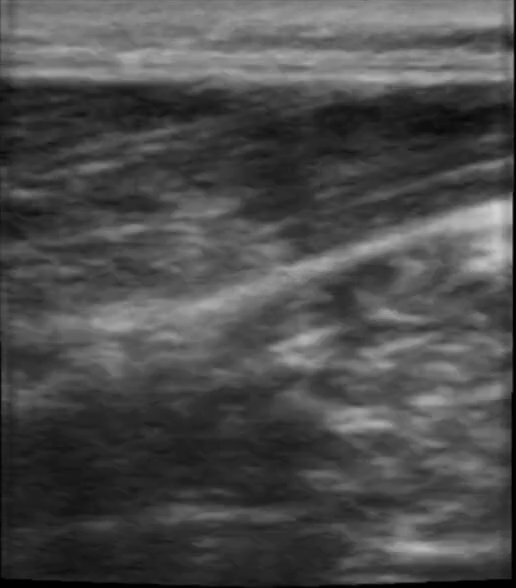} &
    \includegraphics[width=0.25\textwidth]{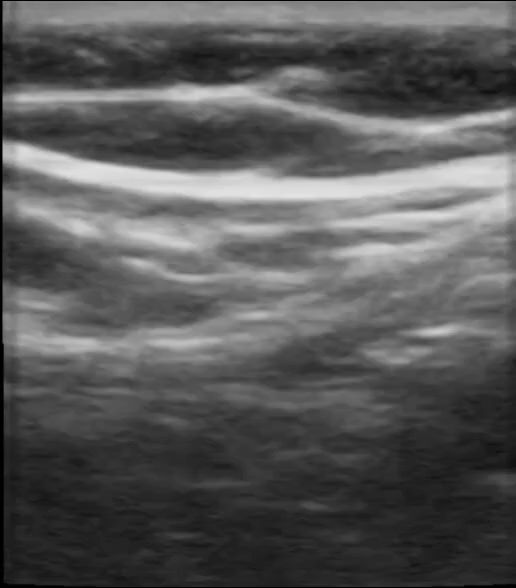} & \includegraphics[width=0.25\textwidth]{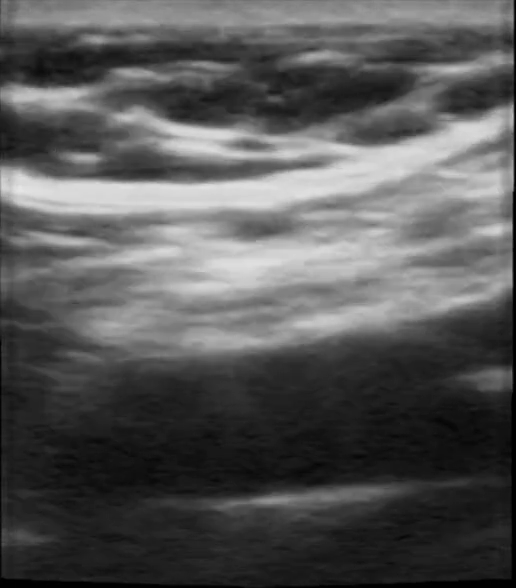} \\ \includegraphics[width=0.25\textwidth]{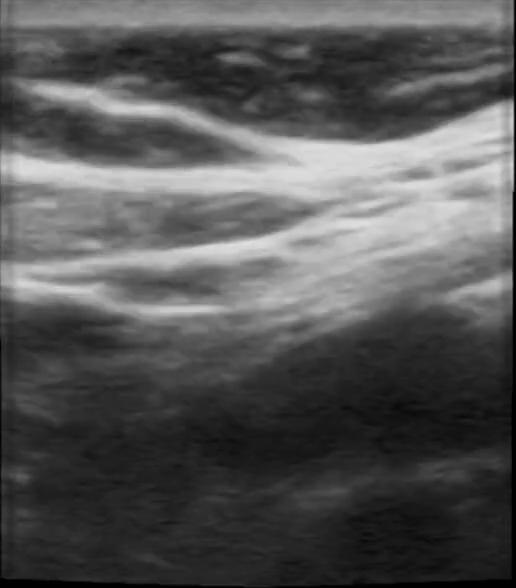} &
    \includegraphics[width=0.25\textwidth]{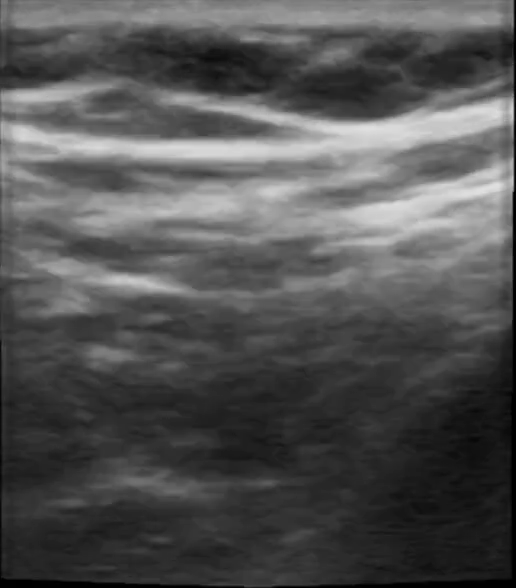} & \includegraphics[width=0.25\textwidth]{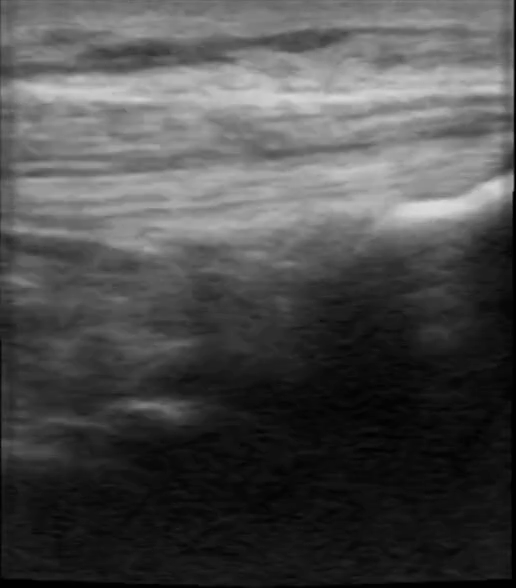} & \includegraphics[width=0.25\textwidth]{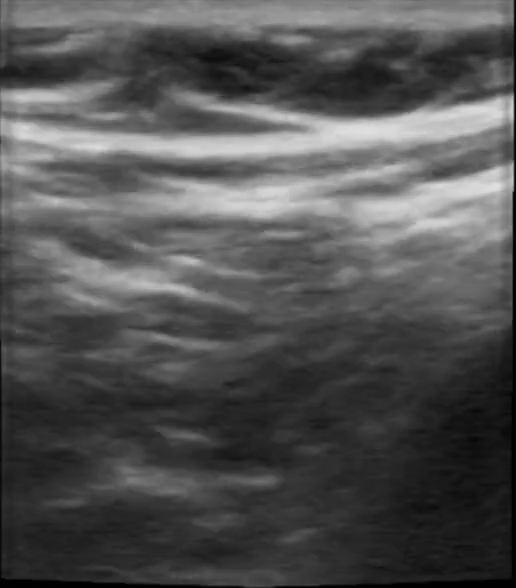}
    \end{tabular}
    \caption{We present additional results using~\acronym. In general, we observe high-quality artifact free reconstructions with our approach.}
    \label{fig:additionalresults}
\end{figure}

\ifdefined\submission
    We present additional qualitative results reconstructed using our method on the Ultrasound in the wild dataset we introduced and show novel views in~\Cref{fig:additionalresults}. Furthermore, we encourage the reader to go through our supplementary for rendered videos with the camera moving around the scene in azimuth with a fixed elevation angle.
\fi
\ifarxivorcameraready
    We present additional qualitative results reconstructed using our method on the Ultrasound in the wild dataset we introduced and show novel views in~\Cref{fig:additionalresults}. Furthermore, we encourage the reader to visit our project website for rendered videos with the camera moving around the scene in azimuth with a fixed elevation angle.
\fi

\section{Ethics Statement}
\label{sec:ethics}

Our approach produces very compelling qualitative and quantitative results for reconstructing ultrasound imaging in the wild. However, in its current form, the authors do not give any theoretical guarantees while using this method and we strongly suggest readers looking to adopt this method to go through the limitations as well. Our data was collected with appropriate institutional guidelines on using medical data from humans. We obtained a waiver from our institution's Research Ethics Board, as all data collected was part of a self study (PNT) for this proof of concept.

\setlength{\tabcolsep}{1pt}


\end{document}